\documentclass[10pt,journal,compsoc]{IEEEtran}
\usepackage{amsmath,amsfonts}
\usepackage{algorithmic}
\usepackage{algorithm}
\usepackage{array}
\usepackage[caption=false,font=normalsize,labelfont=sf,textfont=sf]{subfig}
\usepackage{textcomp}
\usepackage{stfloats}
\usepackage{url}
\usepackage{verbatim}
\usepackage{graphicx}
\usepackage{cite}
\usepackage{scalerel}
\usepackage{color}
\usepackage{color}
\usepackage[english]{babel}
\usepackage{hyperref}
\usepackage{mathtools}
\usepackage{multirow}
\usepackage{makecell}

\title{Semi-sparsity Priors for Image Structure Analysis and Extraction}

\author{Junqing Huang,
	Haihui Wang,
	Michael Ruzhansky
	\thanks{Manuscript received XX, XXXX, 2023; revised XX, XXXX, XX and accepted XX, XXXX, XX. Date of publication XX, XXXX, XX; date of current version XX, XXXX, XX. This work was supported in part by the Research Foundation – Flanders (FWO) Odysseus 1 under Grant G.0H94.18N; Methusalem Programme of the Ghent University Special Research Fund (BOF) under Grant 01M01021; and in part by the National Science and Technology Major Project, China, under Grant J2019-I-0001-0001 and Grant J2019-I-0019-0018. Michael Ruzhansky was also supported by Engineering and Physical Sciences Research Council (EPSRC) under Grant EP/R003025/2. (Corresponding author: Michael Ruzhansky.)}
	\thanks{Junqing Huang and Michael Ruzhansky are with the Department of Mathematics: Analysis, Logic and Discrete Mathematics, Ghent University, 9000 Ghent, Belgium; Michael Ruzhansky is also with the School of Mathematical Sciences, Queen Mary University of London, E1 4NS London, UK (e-mail: \{Junqing.Huang, Michael.Ruzhansky\} @UGent.be).}
	\thanks{Haihui Wang is with the School of Mathematical Sciences, Beihang University (BUAA), China (e-mail: whhmath@buaa.edu.cn).}
    \thanks{Junqing Huang and Haihui Wang contributed equally to this work.}
	\thanks{Digital Object Identifier no. XX.XXXX/TIP.XXXX.XXXXXXX.}}
\begin{document}

\IEEEtitleabstractindextext{
    \begin{abstract}
    
    Image structure-texture decomposition is a long-standing and fundamental problem in both image processing and computer vision fields. In this paper, we propose a generalized semi-sparse regularization framework for image structural analysis and extraction, which allows us to decouple the underlying image structures from complicated textural backgrounds. Combining with different textural analysis models, such a regularization receives favorable properties differing from many traditional methods. We demonstrate that it is not only capable of preserving image structures without introducing notorious staircase artifacts in polynomial-smoothing surfaces but is also applicable for decomposing image textures with strong oscillatory patterns. Moreover, we also introduce an efficient numerical solution based on an alternating direction method of multipliers (ADMM) algorithm, which gives rise to a simple and maneuverable way for image structure-texture decomposition. The versatility of the proposed method is finally verified by a series of experimental results with the capability of producing comparable or superior image decomposition results against cutting-edge methods.  
    
    \end{abstract}

}
\maketitle

\IEEEpubid{0000--0000/00\$00.00~\copyright~2023 IEEE}

\section{Introduction}
\label{sec:introduction}

\IEEEpubidadjcol

\IEEEPARstart{N}atural images always contain various well-organized objects together with some regular/irregular textural patterns, which convey rich information for both machine and human vision. In many psychology and perception~\cite{land1971lightness, gilchrist2006seeing}, it has also found that the contribution of visual cues mostly comes from geometrical structures of objects rather than local individual details. Interestingly, it is always easy for human visual system (HSV) to distinguish main objects from the complex textural backgrounds. Based on these observations, it is always desirable to decouple an image into different meaningful components, for example, a structural part and one or multiple textural parts~\cite{meyer2001oscillating, perona1990scale}. The problem is known as image structure-texture decomposition and has gained great attention in research fields because it helps to extract and analyze the preferred information of images. Such an image decomposition is also proved to be beneficial for a variety of vision-based applications such as image edge  extraction~\cite{meyer2001oscillating, xu2012structure}, cartoon and abstraction~\cite{xu2012structure}, tone mapping and enhancement~\cite{farbman2008edge, ghosh2019fast}, and so on.

\begin{figure}[t]
\centering
{\includegraphics[width=0.42\textwidth]{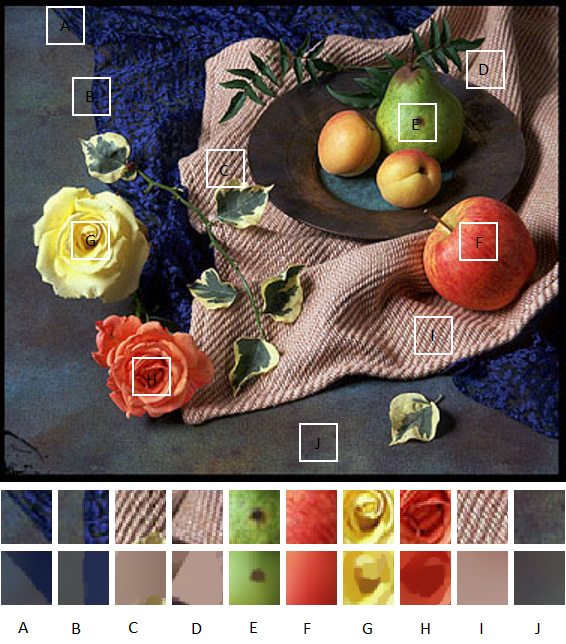}}
\caption{An illustration of image structures and textures with different characteristics in local close-ups (bottom). It is clear that image textures may have various forms with coarse-to-fine oscillating patterns arranged in spatial (non-)uniformly, (ir-)regularly, (an-)isotropically, and so on; while the underlying image structures are in general referred to as strong edges, piecewise constant and/or smoothing surfaces.}
\label{Fig:fig1}
\vspace{-6mm}
\end{figure}

\IEEEpubidadjcol

Decomposing an image into meaningful structural and textural components is generally a challenging problem due to the complex visual analysis processes. As shown in Fig. \ref{Fig:fig1}, visual objects could be exhibited in various and complicated forms --- for example, arranged in spatial (non-)uniformly, (ir-)regularly, (an-)isotropically. The structural and textural parts, as illustrated in~\cite{vese2003modeling, cimpoi2014describing}, could be vague under the semantic descriptions, because they are highly dependent on the scale --- that is, image structures in one scale can be regarded as the textures in another scale. As a result, it is not easy to model them based on a simple model. In practice, it is necessary to characterize both structural and textural parts to make the image decomposition problem to be deterministic when modeling them mathematically.

In a general sense, an image is assumed to be composed of two fundamental parts: a structural part --- referring to the geometrical clues, and a textural part --- representing these coarse-to-fine scale details or noise~\cite{perona1990scale, rudin1992nonlinear, meyer2001oscillating, yin2007total}. It may also suggest that the structural part contains semantic meaning elements such as homogeneous regions, object contours, and sharpening edges, while the textural part could have more abundant and diverse local features, at least the periodic and oscillating information~\cite{meyer2001oscillating, vese2003modeling}. This assumption has been adopted in many existing methods such as partial differential equation (PDE) diffusion methods~\cite{perona1990scale, weickert1998anisotropic}, and structure-aware filtering methods~\cite{farbman2008edge, tomasi1998bilateral, xu2011image}, variational-based methods~\cite{rudin1992nonlinear, vese2003modeling, osher2003image} and higher-order extentions~\cite{bredies2020higher, liu2018new}, and so on. Despite the great success, there is still considerable interest to exploit powerful and efficient methods to achieve visual-appealing decomposition results both theoretically and practically.

In this paper, we propose a novel semi-sparse model for image structure-texture decomposition, which devotes to extracting piece-wise smoothing structures from complex textural backgrounds. This new model is derived from an optimization framework consisting of $L^1$-norm data fidelity and semi-sparse regularization, which takes advantage of two-fold benefits for high-quality decomposition results. On the one hand, it has shown in traditional methods~\cite{duval2009tvl1, yin2005image, chan2005aspects} that $L^1$-norm data fidelity helps to capture the oscillatory patterns and is empirically suitable for structure-texture decomposition. On the other hand, it also turns out in cutting-edge filtering methods~\cite{huang2023semi} that semi-sparsity prior knowledge is beneficial for fitting polynomial-smoothing surfaces while preserving these sparse features such as edges and singularities. As illustrated hereafter, such a combination indeed possesses two-fold benefits of $L^1$-norm data fidelity and semi-sparse regularization and is empirically suitable for image structure-texture decomposition. The main contribution of the paper is summarized as follows: 

\begin{itemize}
\setlength{\itemsep}{0pt}
\setlength{\parsep}{0pt}
\setlength{\parskip}{0pt}
	
\item A simple and effective semi-sparsity minimization model is designed for image structure-texture decomposition, which is an optimization-based framework by means of two-fold benefits: $L^1$-norm data fidelity and semi-sparse regularization.
 
 \item An efficient numerical solution based on a multi-block alternating direction method of multipliers (ADMM) algorithm is introduced to solve the non-convex and non-smooth minimization problem, which gives rise to a powerful tool for image structure-texture decomposition. 
 
 \item A number of experimental results are also presented and discussed to demonstrate its versatility and many benefits to image structure-texture decomposition, where our semi-sparsity model performs more favorable image decomposition results on natural images against many state-of-the-art methods.
		
\end{itemize}

We further point out that the proposed approach has a very simple form and possesses attractive properties in decoupling main structures from complex textures. Despite the non-convex nature of the optimization problem,  the proposed method can be efficiently solved using a multi-block ADMM algorithm and significantly improves the performance of image decomposition, in particular towards natural images. Notice also that the semi-sparsity regularization is also possible to be combined with other functions for a more complex image decomposition --- for example, image structure, texture, and noise decomposition. More details will be discussed below.

The rest paper is organized as follows. In Sec. \ref{sec:strcucture_texture_analysis}, we review the typical methods for modeling the textural and structural parts. In Sec. \ref{sec:proposed_work}, we derive the proposed model in detail and illustrate the properties of $L^1$-norm data fidelity and semi-sparsity regularization. An efficient ADMM numerical solution is then introduced in Sec. \ref{sec:solution}. The performance of the proposed method is compared in Sec. \ref{sec:experimental_results} with a series of experimental results. After that, we give a brief analysis of the possible extensions in  Sec. \ref{extensions_and_analysis}. The conclusion and further work are drawn in Sec. \ref{sec:conclusion}.

\section{Structure and Texture Analysis}
\label{sec:strcucture_texture_analysis}

In the literature, many computational models have been developed to either extract main structures from an image or dedicate an explicit image decomposition. We review some related work regarding structural and textural analysis models for favorable image decomposition results. 

\subsection{Problem Formulation}
\label{subsec:problem_formulation}

We consider an additive form of the image decomposition problem --- that is, given an image $f$, it is possible to form the problem, either explicitly or implicitly, into a general optimization-based framework: 
\begin{equation}
\begin{aligned}
\mathop{\min}_{u,v} \; \mathcal{S}(u)+\mathcal{T}(v), \quad \text{subject to} \quad  u+v=f, 
\end{aligned}
\label{Eq:eq1}
\end{equation}
where $u$ and $v$ are the structural and textural counterparts defined on $\Omega \in \mathbf{R}^2$, typically the same rectangle or a square as that of $f$. The functional $\mathcal{S}(u)$ and $\mathcal{T}(v)$ are defined in some proper spaces to characterize the properties of $v$ and $u$, respectively. In general, $v$ and $v$ could be exhibited in varying and complicated forms in natural images. In practice, $\mathcal{S}(u)$ is advocated to capture the well-structural components such as homogeneous regions and salient edges, and $\mathcal{T}(v)$ may be expected to have the capacity of extracting the textural part such as repeating or oscillating patterns or random noise. The generalization of $\mathcal{S}(u)$ and $\mathcal{T}(v)$ contributes to various existing methods in the literature. 

\subsection{Modeling $\mathcal{S}(u)$ for Structural Analysis} 
\label{subsec:structural_part}

Image ``structure'' is a vague semantic terminology and hard to give a mathematical definition. Intuitively, the structural part contributes to the majority of the semantic information of an image, which is always assumed to be composed of piece-wise constants or smoothing surfaces with discontinuous or sharpening edges. The structural analysis is devoted to finding suitable functions $\mathcal{S}(u)$ to represent image structures. As shown in Tab. \ref{Tab:tab1}, $\mathcal{S}(u)$ is typically formulated as a regularization term in the context of Eq. \ref{Eq:eq1}.

\begin{table*}
	\scriptsize
	\begin{minipage}{0.5\linewidth}
		\caption{Methods of modeling structural parts}
		\label{Tab:tab1}
		\centering
		\setlength{\tabcolsep}{0.3mm}{
			\begin{tabular}{c | c}
			\hline
			Methods & Structural functional $\mathcal{S}(u)$  \;  Description and References\\
			\hline
			\hline
			\multirow{2}{*}{\makecell{Structure-aware\\ filters}}
			& \rule{0pt}{10pt} \quad \qquad \qquad $u = Wf$ \qquad \qquad Local filters~\cite{tomasi1998bilateral,he2012guided, cho2014bilateral} \\
			& \rule{0pt}{10pt} $\mathcal{S}(u)\!\coloneqq\! g(u, \nabla u) \|\nabla u\|^2_{2}$, \quad  Global filters~\cite{farbman2008edge, xu2012structure} \\  
			\hline
			\hline
			\multirow{4}{*}{\makecell{Regularized \\methods}}
			&\rule{0pt}{10pt}  $ \mathcal{S}(u)\!\coloneqq\!\alpha \|\nabla u\|_{1}$,   \qquad TV regularization~\cite{rudin1992nonlinear, yin2005image, meyer2001oscillating, osher2003image}\\ 
			&\rule{0pt}{10pt}  $\mathcal{S}(u)\!\coloneqq\!\alpha \|\nabla u\|_{0}$, \qquad $L_0$ gradient regularization~\cite{xu2011image, ono2017l0}\\ 
			&\rule{0pt}{10pt} $\mathcal{S}(u) \!\coloneqq\!\alpha \|\nabla u \|_{1} \!+\!\beta \|\nabla^2 u \|_{1}$, \quad TV-TV$^2$ models~\cite{bergounioux2010second, papafitsoros2014combined}\\
			&\rule{0pt}{10pt} $\mathcal{S}(u)\!\coloneqq\!\alpha \|\nabla u \!-\! v \|_{1}\!+\!\beta\|\nabla v\|_{1}$, \quad TGV models~\cite{bredies2013properties, bredies2015tgv}\\
			\hline
			\hline
			\multirow{2}{*}{\makecell{Patch-based \\ Models}}
			&\rule{0pt}{10pt} $\nabla_w u:=(u(y)\!-\!u(x)) \sqrt{w(x, y)}$, \quad low-rank~\cite{schaeffer2013low} \\
			&\rule{0pt}{10pt} $\mathcal{S}(u) \!\coloneqq\!\alpha\|\boldsymbol{W} \!\circ\! u\|_1+\beta\|\boldsymbol{L} \!\circ\! \boldsymbol{W} \!\circ\! u\|_1$, \quad non-local~\cite{xu2020cartoon}\\
			\hline
			\hline
		\end{tabular}}
	\end{minipage}
        \hspace{1mm}
	\begin{minipage}{0.5\linewidth}
		\caption{Methods of modeling textural parts}
		\label{Tab:tab2}
		\centering
			\setlength{\tabcolsep}{0.2mm}{
				\begin{tabular}{c|c}
					\hline
					Models & Textural functional $\mathcal{T}(v)$ \quad Description and Reference \\
					\hline
					\hline
					\multirow{4}{*}{\rotatebox[origin=c]{-90}{$f=u+v$}}
					& $\mathcal{T}(v) \!\coloneqq\! \lambda \|v\|^p_{p}$,
					\quad $\|v\|^p_{p} \!\coloneqq\!
					\begin{cases}   
						\|u\!-\!f\|^2_{2}, \; \text{ROF model~\cite{rudin1992nonlinear}} \\
						\|u\!-\!f\|_1, \; \text{$L^1$ space~\cite{alliney1997property, yin2005image}} 
					\end{cases}$ \\  
					& $\mathcal{T}(v) \!\coloneqq\! \lambda\|v\|_{G},    
					 \|v\|_{G}\!\coloneqq\!
					\begin{cases} 
						\inf_g \{\|g\|_{L^{\infty}} \!\mid\!\! v \!\!\in\! G\},\\
						G = \{v \mid v = \operatorname{div}g \},
					\end{cases}$\!\!\!\!\text{$G$ space~\cite{meyer2001oscillating}}\\ 
					&\rule{0pt}{10pt} $\mathcal{T}(v) \!\coloneqq\! \lambda \|v\|_{H^{\!-\!1}}$,  \;
					 $\|v\|_{H^{\!-\!1}}\!\coloneqq\!\|\nabla\!\left(\!\Delta^{-1}\!\right)\! v\|_2^2$,  \text{Hilbert space~\cite{osher2003image}}\\
					\hline
					\hline
					\multirow{2}{*}{\rotatebox[origin=c]{-90}{\!\!\!\!\!$f\!=\!u\!+\!v\!+\!n$}} 
					&\rule{0pt}{15pt} $\lambda \|f\!-\!\!u\!-\!v\!\|_{2}^{2}\!+\!\gamma \|v\|_{G_p}, \!\|v\|_{G_p}\!\!\coloneqq\!\!
					\begin{cases} 
						\inf \{\|g\|_{L^p} \!\!\mid\!\! v \!\in\! G_p\},\\
						G_p \!=\! \{v \!\!\mid\!\! v \!=\! \operatorname{div}g \},
					\end{cases}$\!\!\!\!\!\!\!\!\! $G_p$\text{space~\cite{meyer2001oscillating}}\\
					&\rule{0pt}{15pt} $\lambda \|f\!-\! K\!(u\!+\!v\!)\|_{2}^{2}\!+\!\gamma \|v\|_\ast, 
					\|v\|_\ast \!\coloneqq\!
					\begin{cases}
						v \in  W^{s,p}, \text{Sobolev space~\cite{jung2015simultaneous}}\\
						v \in H^{\!-\!s}, \text{Hilbert space~\cite{lieu2008image}}\\     
					\end{cases}$\\
					\hline
					\hline
			\end{tabular}}
	\end{minipage}
\end{table*}

The early work for structure analysis can be traced back to nonlinear PDE methods for anisotropic diffusion~\cite{meyer2001oscillating, perona1990scale, weickert1998anisotropic}. The diffusion process helps to preserve main image structures and remove local textural details, leading to structure-preserving smoothing effects. This idea is further explored by structure-aware filters for simplification and ease of implementation. The existing filters are generally divided into local ones: bilateral filter~\cite{tomasi1998bilateral}, guided filter~\cite{he2012guided} and many variants,  and global ones: weighted least square (WLS) filter~\cite{farbman2008edge} and  extensions~\cite{min2014fast, kim2017fast}, and so on. Many structure-aware filters such as Laplacian filter~\cite{paris2011local}, and rolling guidance filter (RGF)~\cite{zhang2014rolling} also employ multi-scale analysis techniques to characterize coarse-to-fine structures. Although these filters are not originally designed for the goal of structural analysis, they have been used as practical tools for image decomposition. However, many of them may cause blur effects around strong edges and are not adequate to decouple main structures from strong oscillating textural backgrounds. Recently, some special filters are also designed for image structure extraction such as bilateral texture filter~\cite{cho2014bilateral}, relative total variation (RTV) model~\cite{xu2012structure} and variants~\cite{sun2017image, xu2019structure}, and so on. 

Another trend for image structural analysis is based on variational theory in functional spaces. For example, the pioneering Rudin–Osher–Fatemi (ROF) model~\cite{rudin1992nonlinear} assumes that the structural part $u$ belongs to a bounded variation ($\mathrm{BV}$) space, allowing for piece-wise constant functions to represent image structures. The $\mathrm{BV}$ space assumption of natural images has been extensively studied as a powerful mathematical tool in many computational models since the induced total variational ($\mathrm{TV}$) regularization has a famous structure-preserving property. In theory, it is possible for $\mathrm{TV}$-based methods to capture oscillating signals for better decomposition results in accordance with some concisely-designed textural models --- as interpreted hereafter, $L^1$ space~\cite{yin2005image}, weaker $G$ space~\cite{meyer2001oscillating}, Hilbert space~\cite{osher2003image}, and so on. As demonstrated therein, these spaces allow to decouple the complicated textures from image structures, nevertheless, the $\mathrm{TV}$-based regularization for image structures has two limitations due to the properties of the BV space: (1) the TV-based regularization enforces piece-wise constant results which may cause staircase artifacts, especially in the region of polynomial-smoothing surfaces, and (2) the non-linear thresholding operator of TV-based models give rise to over-smoothing effects around sharpening edges. The defects are further illustrated in the experimental results.

Regarding the staircase artifacts in $\mathrm{TV}$-based models, an effective remedy is based on higher-order methods for image structural analysis~\cite{bergounioux2010second, jung2015simultaneous, papafitsoros2015novel, gao2018infimal}. For example, it turns out in the total generalized variation ($\mathrm{TGV}$)~\cite{bredies2010total} that a higher-order gradient regularization is beneficial for signal/image denoising, as it helps reduce staircase effects substantially when preserving the jump and discontinuities (edges). Based on the property of $\mathrm{TGV}$ regularization, the $\mathrm{TGV}$-$L^1$~\cite{bredies2013properties}, $\mathrm{TGV}$-$\mathrm{OSV}$~\cite{jung2015simultaneous}, $\mathrm{TGV}$-Gabor~\cite{liu2018new} methods have also been proposed to replace the $\mathrm{TV}$-based ones for image structure analysis. The $\mathrm{TV}$-$\mathrm{TV}^2$ model is another higher-order modification based on~\cite{bergounioux2010second}, which advocates a similar property as the $\mathrm{TGV}$-based models with resemble effects. Despite the improvements of these higher-order methods in suppressing the notorious staircase artifacts, they still suffer from over-smoothing problems around the sharpening discontinuous edges~\cite{papafitsoros2015novel, bredies2015tgv}. More recently, a semi-sparsity model~\cite{huang2023semi} is also proposed based on a higher-order regularization model, which is proven to have powerful simultaneous-fitting abilities in both sharpening edges and polynomial-smoothing surfaces. However, it is not suitable for image structure extraction  because of the limitation in large oscillatory textural components.

The efforts of avoiding blurry artifacts around strong edges have also received considerable attention in image structural analysis. The sparsity-inducing $L_0$ regularization has been demonstrated to be effective in preserving strong edges in structure-aware smoothing methods~\cite{xu2011image, ono2017l0}. The advantage is further demonstrated in a higher-order extension for signal/image filtering~\cite{huang2023semi}. However, they are not suitable for image structure-texture decomposition directly due to the limitation of removing high-contrast oscillating details. Recently, a new $L_0$ minimization based on the RTV metric~\cite{xu2012structure} has been proposed for image structure retrieval, which turns out to be, in particular, applicable for image structure extraction~\cite{sun2017image}. Despite the advance of $L_0$ minimization in capturing sharpening edges, it is worth noting that such a minimization problem is a non-convex and non-smooth optimization problem and the numerical solution is computationally expensive in the case of large-scale scenarios, despite the recent endeavor in developing acceleration techniques. Inspired by the properties of $L_0$ minimization~\cite{xu2012structure} and higher-order extension~\cite{huang2023semi}, we will further discuss the properties of $L_0$ norm regularization in higher-order gradient domains and demonstrate its advantages for image structural analysis.

\subsection{Modeling $\mathcal{T}(v)$ for Textures or Noises} 
\label{subsec:textural_part}

The textural part $v$, as aforementioned, can be exhibited in diverse and complicated forms in natural images. In general, image texture is recognized as repeated patterns consisting of oscillating details, random noise, and so on. In many existing methods, the textural part $v$ is modeled as the residual of an image $f$ subtracting the structural part $u$ --- that is, $v \!=\! f \!-\! u$. This simplified assumption has been adopted by many existing methods since it provides an easy but acceptable approximation for modeling image textures mathematically. The typical functional $\mathcal{T}(v)$ for textural analysis in the literature are presented in Tab. \ref{Tab:tab2}.  Accordingly, $\mathcal{T}(v)$ is always treated as a data-fidelity term in the context of Eq. \ref{Eq:eq1}. We briefly discuss their advantages and limitations as follows.

 In the well-known ROF model~\cite{rudin1992nonlinear}, $v$ is simply assumed to be in $L^2$ space with image structures $u$ in $\mathrm{BV}$ space. The ROF model takes advantage of $\mathrm{BV}$ space to penalize oscillating signals when representing the structural part $u$ with piece-wise smoothing functions. However, it is not suitable for large oscillatory textural modes despite using multi-scale analysis~\cite{tadmor2004multiscale}. Along with the ROF model, many image decomposition methods were subsequently proposed for better textural modeling results. A simple but attractive remedy is to model the texture part $v$ in the $L^1$ space instead of the $L_2$, which leads to the well-known $\mathrm{TV}$-$L^1$ model for image textural analysis~\cite{alliney1997property, nikolova2004variational}. The benefits underpin the appealing properties of $L^1$ space that help to decouple large-scale oscillating details, outliers and impulse noise, while preserving geometrical features without eroding main structures~\cite{duval2009tvl1, yin2005image}. Notice that the TV-$L^1$ model needs to solve a type of non-smooth minimization problem and the numerical solution is time-consuming even with a concisely-designed algorithm~\cite{yin2005image, chambolle2011first}.  

It has also witnessed many efforts to find more suitable functions to model the textural part $v$ in the literature. The TV-$G$ model~\cite{meyer2001oscillating}, for example, takes into account a weaker space endowed $G$-norm to replace $L^{p} (p\!\ge\!1)$ space for oscillating patterns. The theoretical justification shows that the $G$ space identifies the Banach space containing signals with large oscillations, in particular image textures and noise. However, a direct solution to the $\mathrm{TV}$-$G$ model is not possible due to the unavailability of the corresponding Euler–Lagrange equation~\cite{vese2003modeling, aujol2005image}. An alternative way is using $G_p (1\!\leq \!p\!\le\!\infty)$ to approximate the problem~\cite{vese2003modeling}. The experimental results also verify its better property for capturing the oscillating patterns~\cite{aujol2006structure}. Based on the $G$-norm space assumption, the model has been further explored and implemented along with different structural analysis models, for example, higher-order $\mathrm{TGV}$ model~\cite{jung2015simultaneous}, edge-driven high-order $\mathrm{BV}$ model~\cite{duan2016edge} and adaptive directional variant~\cite{shi2021image}. Despite the advantages, they may partially decouple image structures into textures and cause blurry effects around strong or sharpening edges.

Similar to the $G$-norm assumption, the oscillating patterns in the Osher-Sole-Vese (OSV) model are measured in $H^{-1}$ space given by the second-order derivative of functions in a homogeneous Sobolev space~\cite{osher2003image}. The numerical solution of the OSV model turns out to be easy to implement based on the corresponding Euler-Lagrange equation. In a case of more generalized theoretical analysis ~\cite{lieu2008image}, it is shown that the component $v$ can be modeled in the Hilbert-Sobolev space $H^{-s}$. Along with the strategy of finding admissible textural spaces, there also exist many other models for texture analysis --- for example, the TV-div(BMO)~\cite{le2005image} and $\mathrm{TV}$-Besov~\cite{garnett2007image}, which are also combined with different regularization terms for better results. In the practical implementations, the use of different spaces for textural analysis has shown limited improvements despite the different properties of these spaces. The phenomenon may be largely due to the complex textural patterns of images, in particular for natural images.

\subsection{Extensive Analysis}

Although we discuss a range of models for structure and texture analysis, there exist many other methods that can be also understood under the interpretation of Eq. \ref{Eq:eq1} explicitly or implicitly. The patch-based models~\cite{schaeffer2013low, ono2014cartoon, xu2020cartoon}, for example, characterize a low-rank nature of image structures or textures over non-local image patches. It is possible for them to produce better decomposition results but the performance is at the expense of searching local similar patches. Besides, the frequency filters, transformed domain analysis~\cite{chen2020l0}, and sparse representation techniques~\cite{wang2022structure} are also proposed for structural and textural analysis, while the performance is generally limited in some typical cases and not universally applicable in many practical applications. In addition, it has also been witnessed that some existing methods~\cite{moreno2015adaptive} strive to decompose an image into more than two basic components --- for example, structures, textures, and random noise. Despite their different forms, they share a similar assumption of image structures and textures as listed in Tab. \ref{Tab:tab1} and \ref{Tab:tab2}. More detailed discussions of them are out of the scope here. The reader is also referred to the related research points for more theoretical analysis and their internal connections. It is worth noting that it is still a fertile research point to explore more advanced models for image structure and texture analysis.

\begin{figure*}[!t]
    \begin{center}
     {\includegraphics[width = \textwidth]{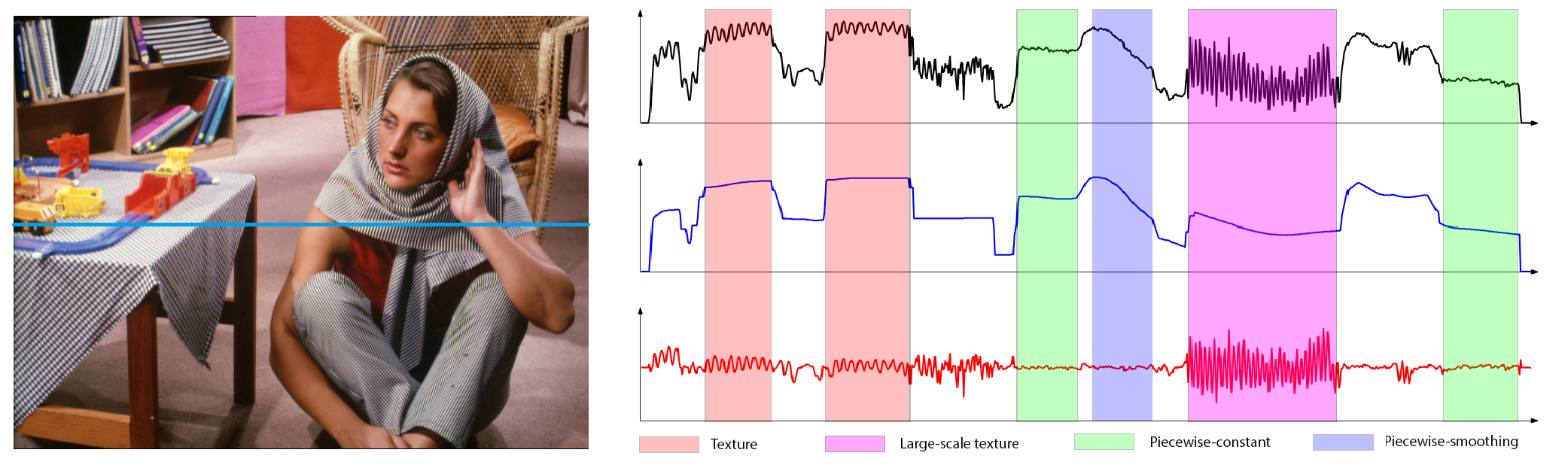}}
    \end{center}
    \caption{An illustration of the characteristics of natural image structures and textures. The input image signal (Black) along a row is plotted with the decomposed structures (\textcolor{blue}{Blue}) and textures (\textcolor{red}{Red}) based on the proposed method. It is clear that the structural part consists of some strong edges, piece-wise constant and polynomial smoothing surfaces, while the textural part contains different levels of oscillation or repeated details.} 
    \label{Fig:fig2}
\end{figure*}

\section{Proposed Method}
\label{sec:proposed_work}

In this section, we further analyze the characteristics of image structures and textures and propose a generalized framework for image structure-texture decomposition. The model is based on a semi-sparsity-inducing regularization for structural analysis. We interpret that such a model is preferable to produce better image decomposition results in accordance with different textural analysis models.

\subsection{Observations and Motivations}
\label{subsec:observation}

We have discussed several established image structure and texture analysis methods and highlighted their strengths and limitations for image decomposition applications. To further elucidate their properties, we first take into account an example in \ref{Fig:fig2}, where a natural image is given with the structural component formed by singularities, sharpening edges, polynomial-smoothed surfaces, and the textural component composed of various characteristics spanning from textural patterns, scales, and directions to amplitudes.

It is always expected for an image decomposition to ideally distinguish the structural and textural parts. But, it is a challenging problem due to the facts: (1) the sharpening edges (singularities) and oscillation patterns are highly intertwined for their high-frequent nature, but the former always belongs to image structures, and the latter is mostly attributed to image textures, (2) the piece-wise constant and polynomial smoothing surfaces may coexist in image structures, but they have very different properties and it is difficult for a simple model to capture their characteristics simultaneously, and (3) the textural part may be exhibited in various forms which are not adequate for a simple model to identify them mathematically. In summary, a plausible image decomposition method should be able to distinguish the three-fold characteristics of signals:

\begin{itemize}
\setlength{\itemsep}{0pt}
\setlength{\parsep}{0pt}
\setlength{\parskip}{0pt}
	
	\item \textbf{Piece-wise constant or smoothing surfaces}: Many natural or synthesized images  have visual objects with homogeneous regions consisting of piece-wise constant and smoothing surfaces simultaneously. It is necessary for image decomposition methods to have the capacity of representing and extracting them without introducing artifacts such as staircase results in the polynomial-smoothing surfaces. 
	
	\item \textbf{Sharpening edges/singularities}: The  discontinuous boundaries between homogeneous objects give rise to sharpening edges or singularities of an image, which is important for visual understanding and analysis. An effective image decomposition model should be able to precisely preserve these features without introducing blur artifacts, and distinguish these sharpening edges and singularities from large oscillating textures\footnote{The assignment of singularities may depend on the appropriate semantic scale of the interested visual information.}. 
	
	\item \textbf{Coarse-to-fine oscillatory details:} The characteristics of oscillatory details can be complicated because of the diverse patterns, multiple directions, varying amplitudes, and coarse-to-fine oscillation scales. It is crucial for an image decomposition method to capture these various  characteristics and distinguish them from sharpening edges and singularities.
	
\end{itemize}

As verified in Fig. \ref{Fig:fig2}, the key point of image structure-texture decomposition is to distinguish the three types of signals while maintaining a fine balance. The dilemma also exists in many existing methods in Tab. \ref{Tab:tab1} and Tab. \ref{Tab:tab2}. Structure-aware filters, for example, are capable of keeping main structures but they do not well distinguish sharpening edges and large oscillating textures. The TV-based regularization is proposed to identify the structural part with piece-wise constant and sharpening edges, while it may result in stair-case artifacts in polynomial-smoothing surfaces~\cite{rudin1992nonlinear, meyer2001oscillating, vese2003modeling}. Higher-order TV extensions~\cite{bergounioux2010second, jung2015simultaneous} alleviate the stair-case artifacts, but they cannot get rid of blurry edges. The textural analysis models perform better textural results with appropriate structural models but they may still have problems in decoupling the coarse-to-fine textures. The $L^1$~\cite{alliney1997property, nikolova2004variational}, $G$-norm~\cite{meyer2001oscillating}, and $H^{-1}$~\cite{osher2003image} spaces help to characterize large-scale oscillating textures, while the performance is somehow still limited in complicated textural patterns and they may still cause side-effects such as over-smoothing results around strong edges.

\begin{figure*}[!t]
	\begin{center}
		\subfloat[Input]
		{\includegraphics[width=0.16\textwidth]{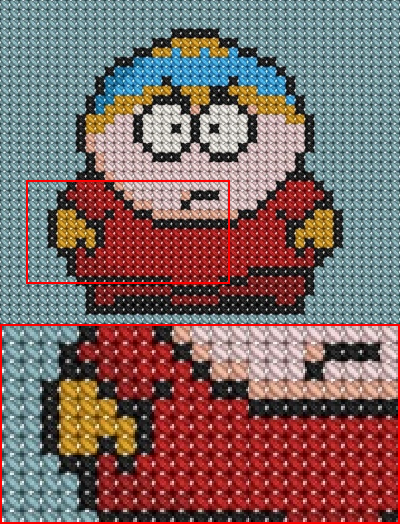}}\hfil
		\subfloat[ROF~\cite{rudin1992nonlinear}]
		{\includegraphics[width=0.16\textwidth]{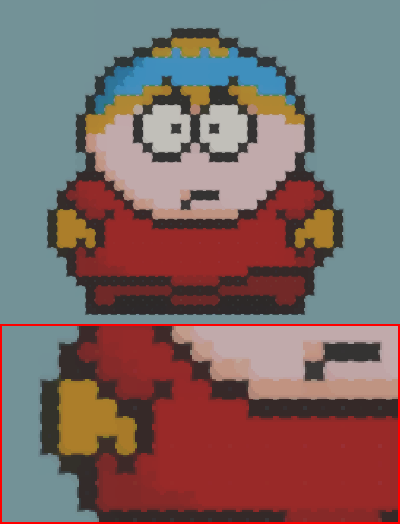}}\hfil
		\subfloat[$\mathrm{TV}$-$L^1$~\cite{le2014cartoon+}]
		{\includegraphics[width=0.16\textwidth]{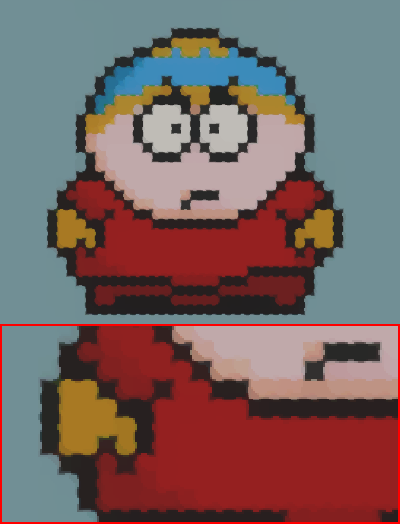}}\hfil
		\subfloat[$\mathrm{TV}$-$G$~\cite{meyer2001oscillating}]
		{\includegraphics[width=0.16\textwidth]{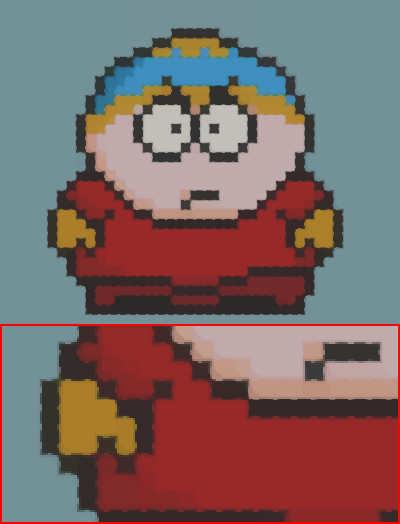}}\hfil
		\subfloat[$\mathrm{TV}$-$\mathcal{H}$~\cite{osher2003image}]
		{\includegraphics[width=0.16\textwidth]{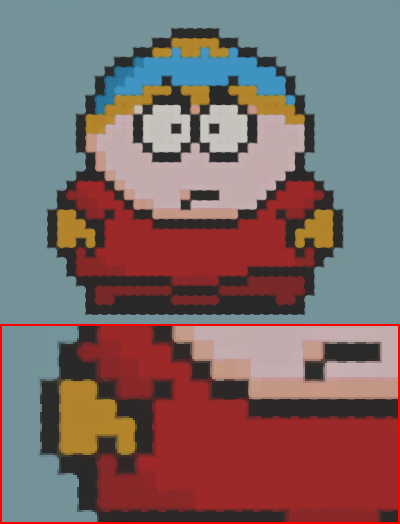}}\hfil
		\subfloat[$\mathrm{TV}$-$G$-$\mathcal{H}$~\cite{xu2022new}]
		{\includegraphics[width=0.16\textwidth]{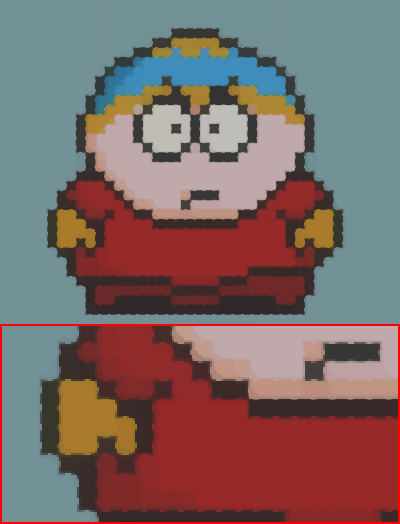}}\hfil
		\subfloat[BTF~\cite{cho2014bilateral}]
		{\includegraphics[width=0.16\textwidth]{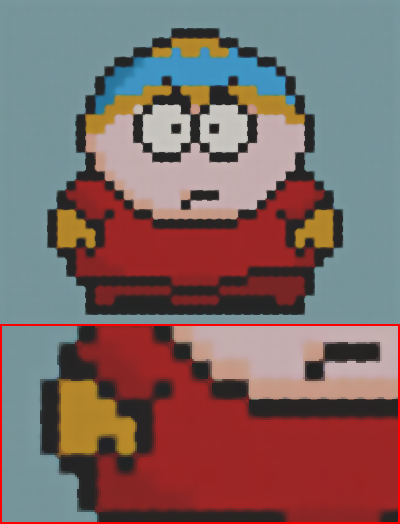}}\hfil
		\subfloat[RTV~\cite{xu2012structure}]
		{\includegraphics[width=0.16\textwidth]{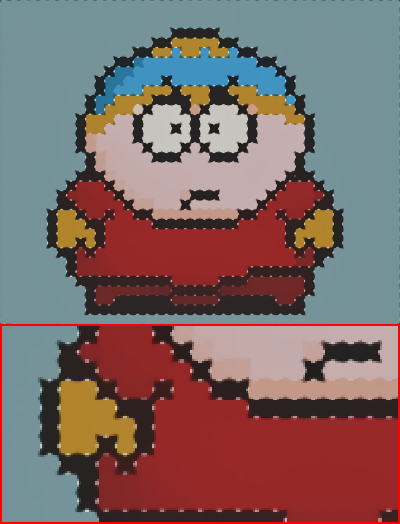}}\hfil
		\subfloat[$\mathrm{TGV}$-$L^1$~\cite{bredies2013properties}]
		{\includegraphics[width=0.16\textwidth]{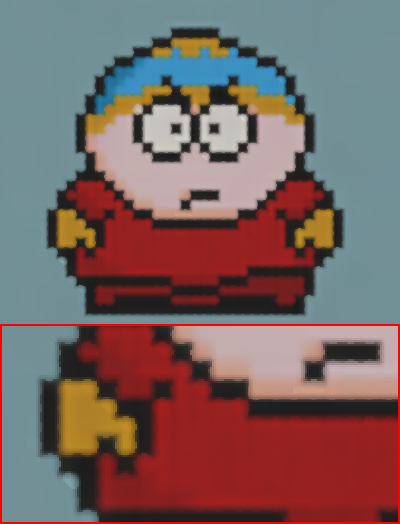}}\hfil
		\subfloat[$\mathrm{TGV}$-$\mathcal{H}$~\cite{jung2015simultaneous}]
		{\includegraphics[width=0.16\textwidth]{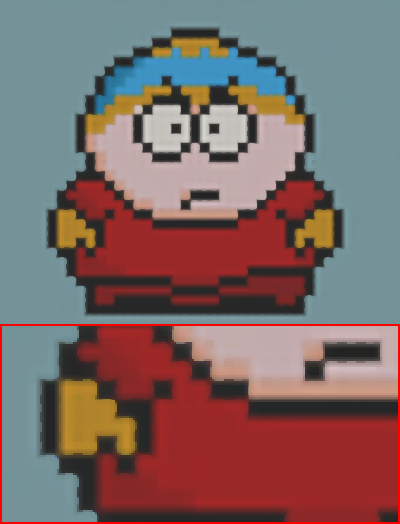}}\hfil
		\subfloat[$\mathrm{HTV}$-$\mathcal{H}$~\cite{jung2015simultaneous}]
		{\includegraphics[width=0.16\textwidth]{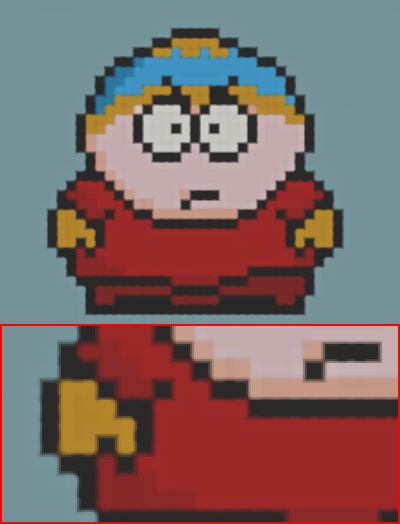}}\hfil
		\subfloat[Ours]
		{\includegraphics[width=0.16\textwidth]{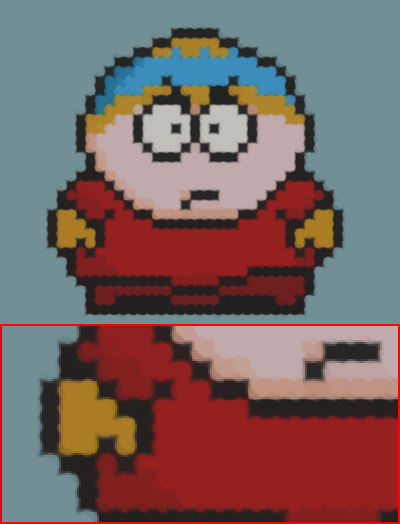}}\hfil
	\end{center}
	\caption{Visual results of image structures in case of \textbf{piece-wise constant structures} and \textbf{large oscillating textures}. (a) Input,  (b) ROF ($\lambda\!=\!0.005, \alpha\!=\!0.001$)~\cite{rudin1992nonlinear},  (c) $\mathrm{TV}$-$L^1$ ($\lambda\!=\!0.003, \alpha\!=\!0.0045$)~\cite{le2014cartoon+},  (d) $\mathrm{TV}$-$G$ ($\lambda\!=\!0.003, \alpha\!=\!0.0005, \gamma\!=\!0.0002$)~\cite{meyer2001oscillating},  (e) $\mathrm{TV}$-$\mathcal{H}$ ($\lambda\!=\!0.002, \alpha\!=\!0.005$)~\cite{osher2003image},  (f) $\mathrm{TV}$-$G$-$\mathcal{H}$ ($\lambda\!=\!0.004, \alpha\!=\!0.001, \gamma\!=\!0.002$)~\cite{xu2022new},  (g) BTF ($\sigma\!=\!5.0, iter\!=\!4$)~\cite{cho2014bilateral},  (h) RTV ($\lambda\!=\!0.01, \sigma\!=\!3.0$)~\cite{xu2012structure},  (i) $\mathrm{TGV}$-$L^1$ ($\lambda\!=\!0.0007, \alpha\!=\!0.0008, \beta\!=\!0.0008$)~\cite{bredies2013properties},  (j) $\mathrm{TGV}$-$\mathcal{H}$ ($\lambda\!=\!0.004, \alpha\!=\!0.01, \beta\!=\!0.03$)~\cite{jung2015simultaneous},  (k) $\mathrm{HTV}$-$\mathcal{H}$ ($\lambda\!=\!0.003, \alpha\!=\!0.006, \beta\!=\!0.0015$)~\cite{jung2015simultaneous}, and (l) Ours ($\lambda\!=\!0.005, \alpha\!=\!0.006, \beta\!=\!0.001$). For fairness, all methods are fine-tuned for a similar level of smoothness. $SSR (C_0/C_1)$, (b)$\scriptsize{\sim}$(j):17.24 (0.1138/0.0194),	17.16 (0.0301/0.0340),	17.15 (0.1035/0.0075),	17.18 (0.0364/0.0316),	17.11 (0.1298/0.0059),	17.05 (0.0369/0.0727),	17.17 (0.0427/0.1089),	17.04 (0.0645/0.0609),	17.06 (0.0717/0.0627),	17.07 (0.0560/0.0829),	17.12 (0.0203/0.0022). (Zoom in for better view.)} 
    \label{Fig:fig3}
\end{figure*} %

\subsection{(Semi)-Sparsity Inducing Regularization}

``Sparsity'' prior knowledge has been extensively studied in many signal and image processing fields, for example, sparse image recovering~\cite{mairal2014sparse} sparse representation~\cite{elad2010sparse}, compressed sensing~\cite{donoho2006compressed}, and so on. It has shown that $L_0$ norm regularization can be used as a favorable and powerful mathematical tool to identify sparsity-induced priors because of the capacity in capturing the minority of key features (archetypes) of signals. In many linear system backgrounds, ``sparsity'' priors can be posed into a $L_0$ quasi-norm regularized optimization model with the form,
\begin{equation}
	\begin{aligned}
		\mathop{\min}_{u} \Phi(u)+\alpha {\left\Vert Hu \right\Vert }_0,
	\end{aligned} 
	\label{Eq:eq2}
\end{equation}
where $\Phi(u)$ is defined in a proper function space modeling the forward process of a physical system. The second term provides a ``sparsity'' constraint under a linear operator $H$ with a positive weight $\alpha$. The notation ${\left\Vert \cdot\right\Vert }_0$ is the so-called $L_{0}$ quasi-norm denoting the number of non-zero entries of a vector, which provides a simple and easily-grasped measurement of sparsity. 

Many sparsity-inducing models can be understood in the context of Eq. \ref{Eq:eq2}. The inverse imaging reconstruction, for example, takes into account $\Phi(u)\!=\!{\left\Vert Au\!-\!f \right\Vert }_2^2$ where $A$ is observation matrix and $H=\nabla$ is the gradient operator because the $L_0$ norm gradient regularization help to restore piece-wise constant image surfaces. This idea has also drawn great attention in some filtering techniques~\cite{xu2011image, ono2017l0} because it helps to preserve the minority of singularities and discontinuous sharpening features of signals. Recently, such an idea is also found in image structure restoration~\cite{sun2017image}, where  a sparsity-induced regularization is introduced for image structure analysis. 

Despite the progress in preserving discontinuous and sharp edges, the occurrence of notorious stair-case artifacts is not negligible in existing sparsity-inducing models. To address this contradiction, there has been considerable attention towards using regularization techniques in higher-order gradient domains, in particular focusing on $\mathrm{TGV}$-based methods~\cite{bredies2010total, jung2015simultaneous} and $\mathrm{TV}$-$\mathrm{TV}^2$-based regularization~\cite{bergounioux2010second, jung2015simultaneous}. More recently, a so-called semi-sparsity model has been also explored for smoothing filters~\cite{huang2023semi, bredies2010total}, in which the established $L_0$ regularization in higher-order gradient domains allows simultaneously fitting the regions coexisting sharp edges and polynomial-smoothing surfaces. Motivated by the edge-preserving property of $L_0$ regularization \cite{xu2011image, ono2017l0} and higher-order methods~\cite{huang2023semi,bredies2010total} in avoiding stair-case artifacts,  we combine them in a unified framework and illustrate that they are also universally applicable for image structural and textural analysis. 

Formally, we recall the semi-sparse notations and show how to identify the semi-sparse property in the higher-order gradient domains effectively. Without loss of the generality~\cite{huang2023semi}, we say $u$ is a semi-sparse signal if its higher-order gradients satisfy the following relationship, 
\begin{equation}
\small
\begin{aligned}
\begin{cases}
{\left\Vert {\nabla}^{n\!-\!1} u\right\Vert }_0>N,\\
{\left\Vert {\nabla}^{n} u\right\Vert }_0<M,
\end{cases}
\end{aligned}
\label{Eq:eq3}
\end{equation}
where ${\nabla}^{n}$ is the $n$-th (partial) differential operator, and $M, N$ are some appropriate natural number satisfying ${N\gg M}$. The merit of Eq. \ref{Eq:eq3} is easy to understand, that is, the non-zeros entries of ${\nabla}^{n} u$, measured by $L_{0}$ quasi-norm, is much smaller than that of ${\nabla}^{n\!-\!1} u$, which occurs if and only if ${\nabla}^{n} u$ is much more sparse than ${\nabla}^{n\!-\!1} u$. Taking into account a $n$ degree piece-wise polynomial function for example, it is easy to verify that the $n$-th order gradient ${\nabla}^{n} u$ is sparse, while it not holds for the $k$-th $(k<n)$ order gradient ${\nabla}^{k} u$. It is also clear that a polynomial-smoothing signal with degree $d<n$  satisfies Eq. \ref{Eq:eq3} essentially. 

\begin{figure*}[!t]
	\begin{center}
		\subfloat[Input]
		{\includegraphics[width=0.16\textwidth]{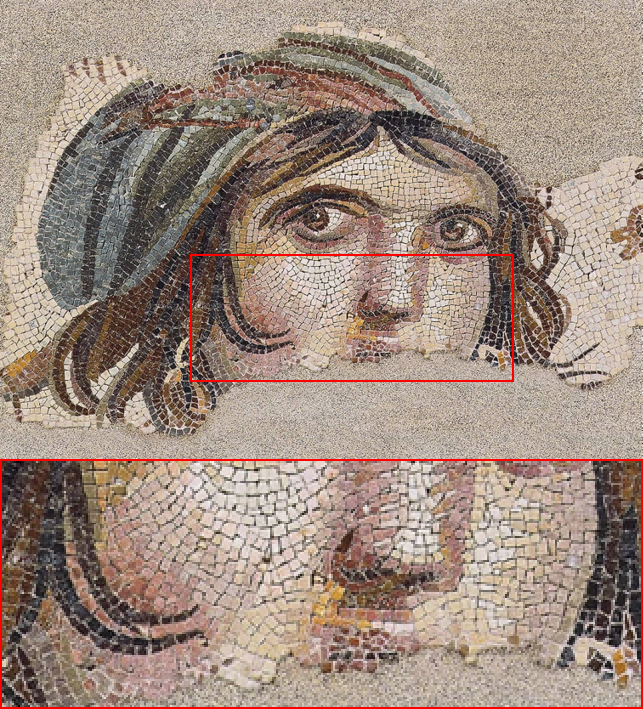}}\hfil
		\subfloat[ROF~\cite{rudin1992nonlinear}]
		{\includegraphics[width=0.16\textwidth]{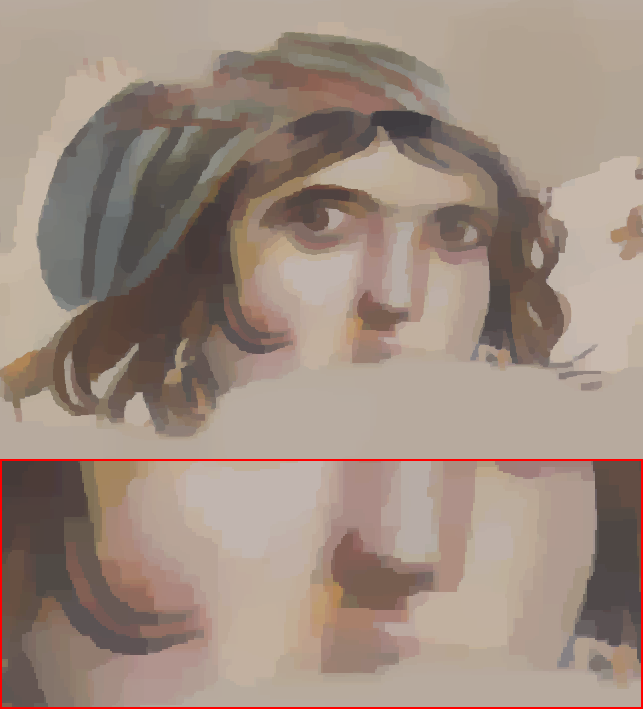}}\hfil
		\subfloat[$\mathrm{TV}$-$L^1$~\cite{le2014cartoon+}]
		{\includegraphics[width=0.16\textwidth]{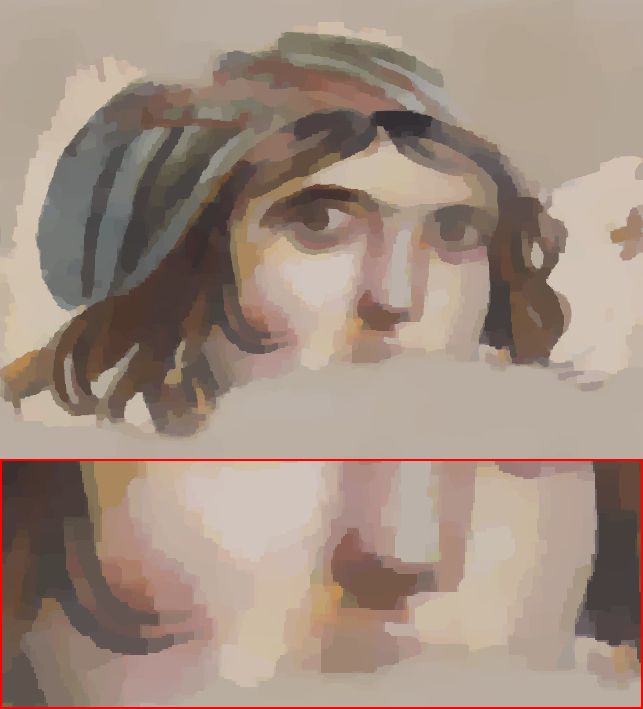}}\hfil
		\subfloat[$\mathrm{TV}$-$G$~\cite{meyer2001oscillating}]
		{\includegraphics[width=0.16\textwidth]{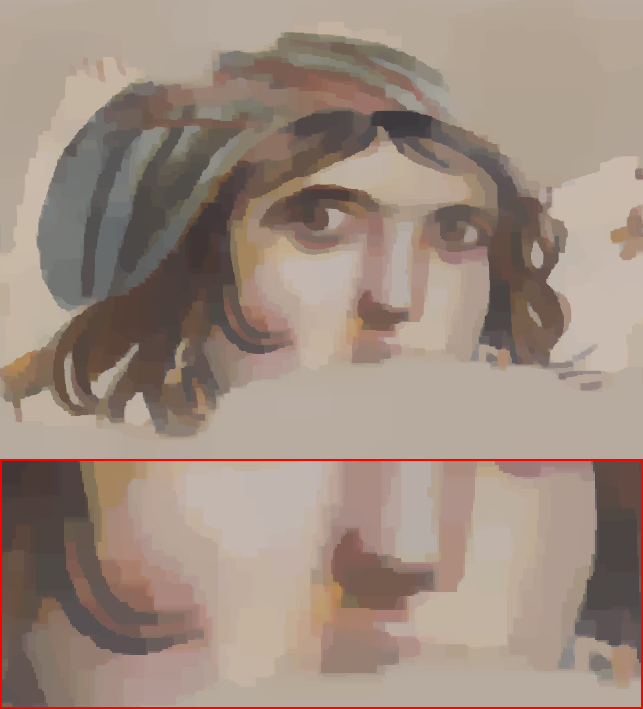}}\hfil
		\subfloat[$\mathrm{TV}$-$\mathcal{H}$~\cite{osher2003image}]
		{\includegraphics[width=0.16\textwidth]{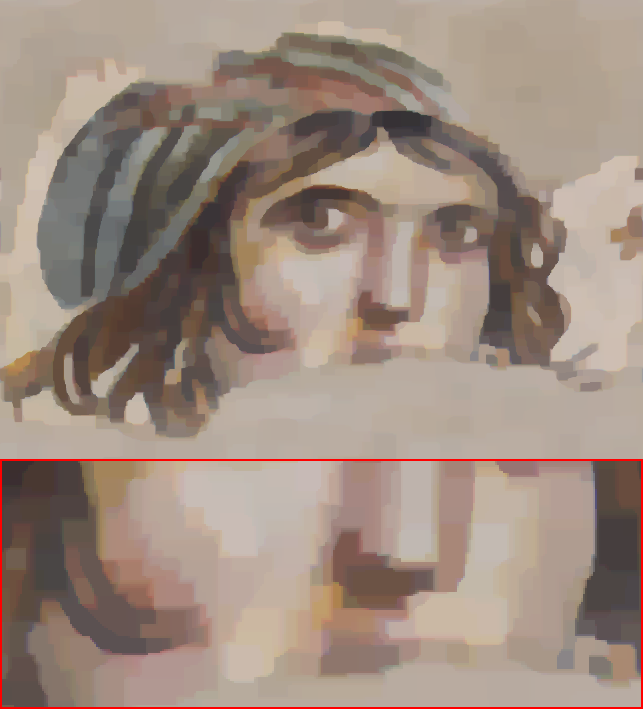}}\hfil
		\subfloat[$\mathrm{TV}$-$G$-$\mathcal{H}$~\cite{xu2022new}]
		{\includegraphics[width=0.16\textwidth]{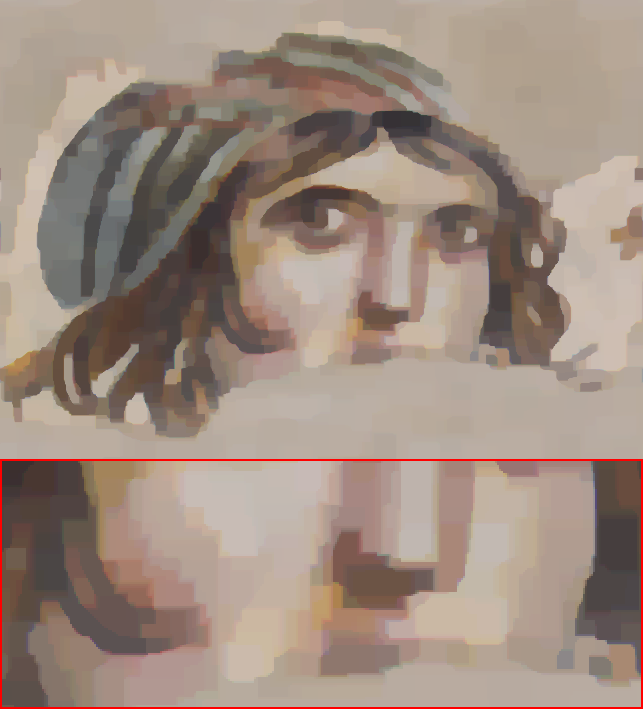}}\hfil
		\subfloat[BTF~\cite{cho2014bilateral}]
		{\includegraphics[width=0.16\textwidth]{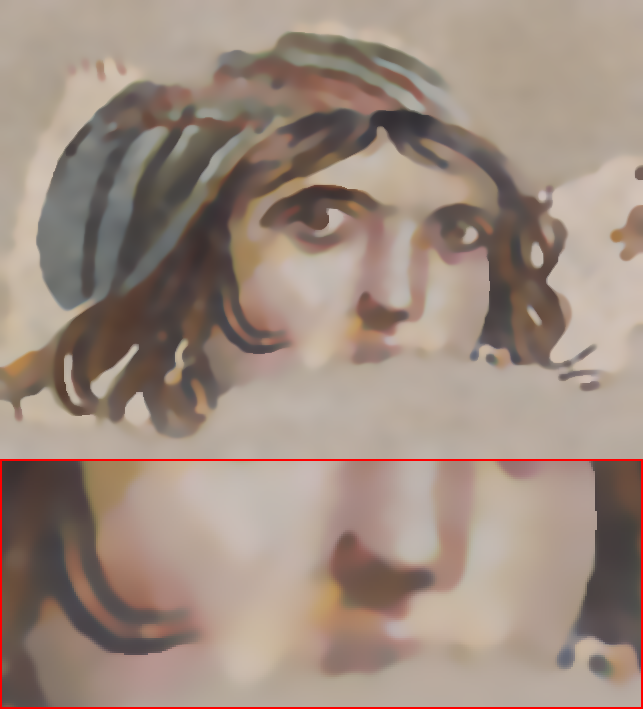}}\hfil
		\subfloat[RTV~\cite{xu2012structure}]
		{\includegraphics[width=0.16\textwidth]{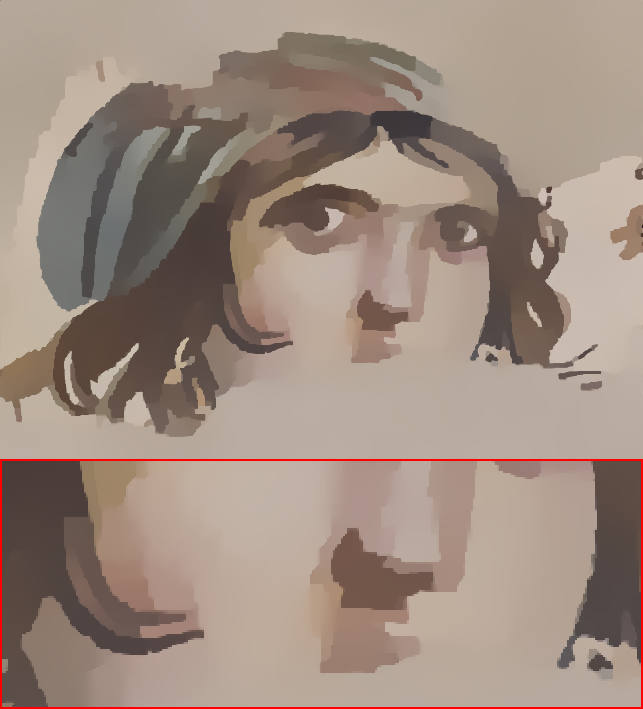}}\hfil
		\subfloat[$\mathrm{TGV}$-$L^1$~\cite{bredies2013properties}]
		{\includegraphics[width=0.16\textwidth]{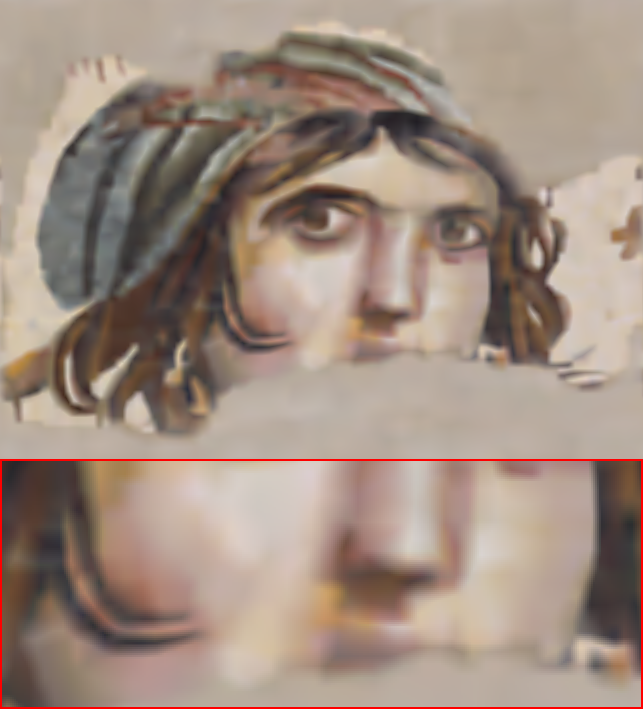}}\hfil
		\subfloat[$\mathrm{TGV}$-$\mathcal{H}$~\cite{jung2015simultaneous}]
		{\includegraphics[width=0.16\textwidth]{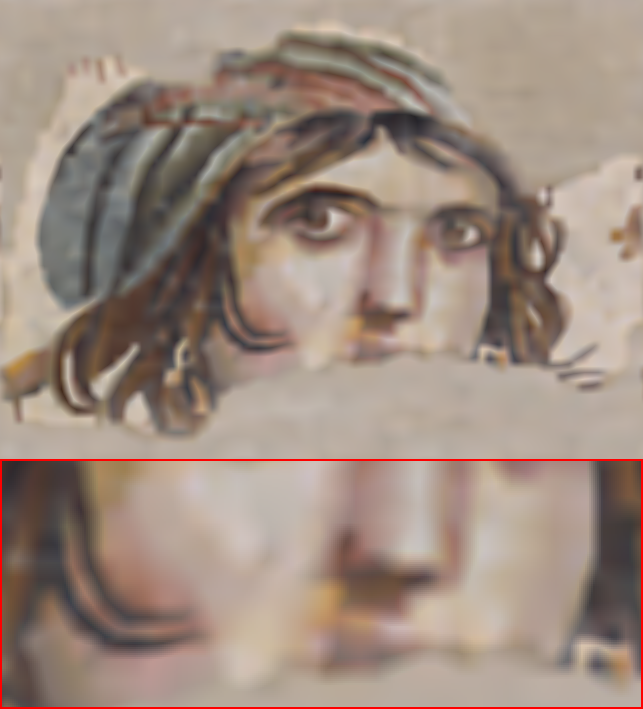}}\hfil
		\subfloat[$\mathrm{HTV}$-$\mathcal{H}$~\cite{jung2015simultaneous}]
		{\includegraphics[width=0.16\textwidth]{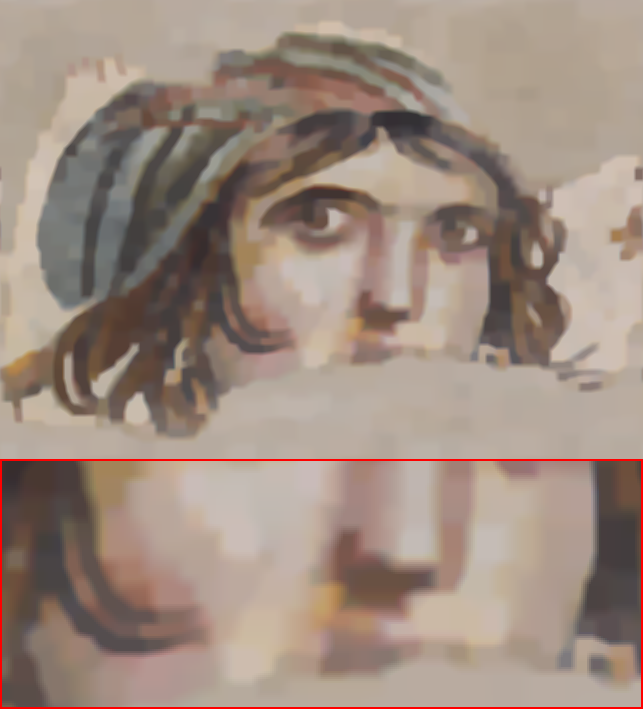}}\hfil
		\subfloat[Ours]
		{\includegraphics[width=0.16\textwidth]{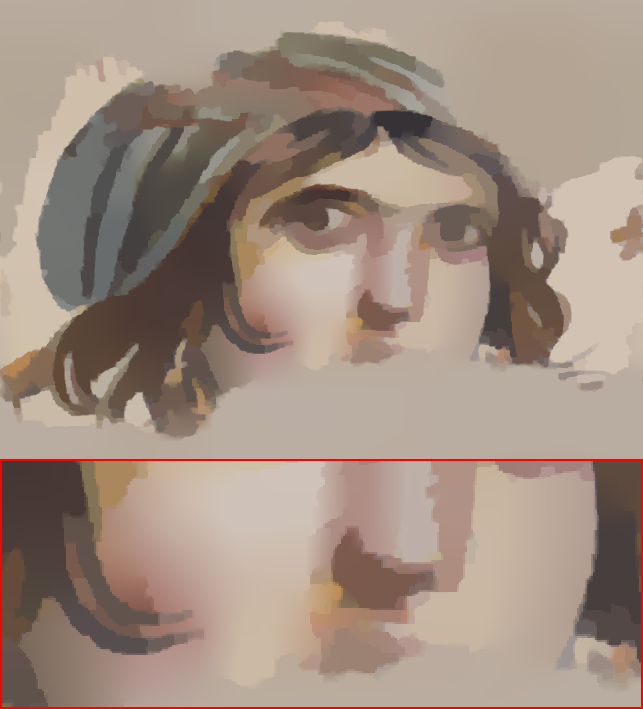}}\hfil
	\end{center}
	\caption{Visual results of image structures in case of \textbf{piece-wise constant/smoothing structures} and \textbf{large oscillating textures}. (a) Input,  (b) ROF ($\lambda\!=\!0.0035$, $ \alpha\!=\!0.001$)~\cite{rudin1992nonlinear},  (c) $\mathrm{TV}$-$L^1$ ($\lambda\!=\!0.001$, $\alpha\!=\!0.002$)~\cite{le2014cartoon+},  (d) $\mathrm{TV}$-$G$ ($\lambda\!=\!0.003$, $\alpha\!=\!0.0007$, $\gamma\!=\!0.0002$)~\cite{meyer2001oscillating},  (e) $\mathrm{TV}$-$\mathcal{H}$ ($\lambda\!=\!0.002$, $\alpha\!=\!0.003$)~\cite{osher2003image},  (f) $\mathrm{TV}$-$G$-$\mathcal{H}$ ($\lambda\!=\!0.004$, $\alpha\!=\!0.008$, $\gamma\!=\!0.008$)~\cite{xu2022new}, (g) BTF ($\sigma\!=\!5.0$, $ iters\!=\!7$)~\cite{cho2014bilateral}, (h) RTV ($\lambda\!=\!0.015$, $ \sigma\!=\!4.0$)~\cite{xu2012structure}, (i) $\mathrm{TGV}$-$L^1$ ($\lambda\!=\!0.001$, $ \alpha\!=\!0.002$, $\beta\!=\!0.002$)~\cite{bredies2013properties},  (j) $\mathrm{TGV}$-$\mathcal{H}$ ($\lambda\!=\!0.002$, $ \alpha\!=\!0.05$, $\beta\!=\!0.04$)~\cite{jung2015simultaneous}, (k) $\mathrm{HTV}$-$\mathcal{H}$ ($\lambda\!=\!0.002$, $\alpha\!=\!0.0045$, $\beta\!=\!0.004$)~\cite{jung2015simultaneous}, and (l) Ours ($\lambda\!=\!0.004$, $ \alpha\!=\!0.007$, $\beta\!=\!0.002$). Quantitative results with $STR (C_0/C_1)$ metrics, (b)$\scriptsize{\sim}$(j):
    18.88(0.1220/0.0158),	18.89(0.0365/0.0204),	18.90(0.1101/0.0259),	18.89(0.0486/0.0429),	18.84(0.0444/0.0500),	19.01(0.0373/0.0834),	18.97(0.0539/0.0195),	18.67(0.1003/0.1742),	18.86(0.0320/0.2002),	18.87(0.0431/0.1421),	18.95(0.0292/0.0188). (Zoom in for better view.)}
    \label{Fig:fig4}
\end{figure*} %

\subsection{Generalized Semi-Sparse Decomposition}

It has shown in the work~\cite{huang2023semi} that natural images always exhibit notable semi-sparsity properties when attributing the characteristics of higher-order gradient distributions. Notice that image structures could characterize more obvious semi-sparsity properties, as they can be effectively represented as a combination of a series of piece-wise constant regions, polynomial-smoothing surfaces, and sharpening discontinuous edges. This observation motivates us to use semi-sparsity priors to capture structural features and avoid blurry and stair-case artifacts. Specifically, we propose a general semi-sparsity framework based on the $L_0$ regularization in higher-order gradient domains, 
\begin{equation}
\small
\begin{aligned}
\mathop{\min}_{u,v} \mathcal{T}(v)\!+\!\alpha_{k} \sum_{k=1}^{n-1}  {\left\Vert {\nabla}^{k}u\right\Vert }_1\!+\!\alpha_{n}{\left\Vert {\nabla}^{n} u \right\Vert }_0, \quad \textit{s.t.} \quad u+v =f,
\end{aligned}
\label{Eq:eq4}
\end{equation}
where $\alpha=[\alpha_1, \alpha_2, \cdots, \alpha_n]$ are the positive weights. The semi-sparse regularization part advocates that the $k$-th ($k\!<\!n$) order gradients ${ {\nabla}^{k} u}$ have a small measure in $L^1$ space, and the highest-order gradient ${ {\nabla}^{n} u}$ tends to be fully sparse in views of the polynomial-smoothing property of image structures. Differing from the semi-sparsity smoothing~\cite{huang2023semi}, a new semi-sparse regularization is introduced here \footnote{In the semi-sparsity smoothing model~\cite{huang2023semi} the $k$-th ($k\!<\!n$) order gradients ${ {\nabla}^{k} u}$ is assumed to be close to that of image $f$ in $L^2$ space.}. As we  illustrate hereafter, such a semi-sparse regularization combined with an appropriate structural analysis model $\mathcal{T}(v)$ enjoys better fitting performance in preserving polynomial-smoothing surfaces and sharp edges.

In principle, $\mathcal{T}(v)$ in Eq. \ref{Eq:eq4} can be arbitrarily chosen from Tab. \ref{Tab:tab2}.  However, the textural part $v$, as illustrated in Fig. \ref{Fig:fig2}, varies from image to image with diverse and multiple forms spanning from textural patterns, scales, directions to amplitudes, and so on. It is usually difficult to discriminate and describe them independently in a simple model. In this paper, we suggest using the $L^1$-norm to model image textures due to its simplicity and effectiveness. Firstly, it has been demonstrated in several TV-$L^1$ models~\cite{alliney1997property, yin2005image} that $L^1$ space allows representing image textures with a variety of patterns, including coarse-to-fine scales, anisotropic directions, varying oscillating amplitudes frequencies, and so on. Secondly, the contrast of image structures is also important information that needs to be preserved during the decomposition process, however, it is not adequate for many decomposition methods such as the $L_2$-norm model. Instead, $\mathcal{T}(v)$ in $L^1$ space has been proven to have the capacity of preserving the contrast information even in large oscillating textural cases. Moreover, the $L^1$ data fidelity provides a fine balance between effectiveness and efficiency due to the simple form and numerical implementation, and affordable computational  cost. As demonstrated in experimental results, such a semi-sparse regularization combined with $L^1$
data-fidelity enjoys better fitting performance in polynomial-smoothing surfaces and sharpening edges.

Furthermore, the choice of $ n $ is also an important factor for the image composition model of Eq. \ref{Eq:eq4}, which is in nature determined by the property of image structures. For natural images, it has been shown in~\cite{huang2023semi} that $n=2$ is enough to give plausible filtering results in practice. Recalling the similar properties of image structures and the observations in Sec. \ref{subsec:observation}, we here choose $n=2$ as the highest order for semi-sparsity regularization, reducing Eq. \ref{Eq:eq4} into the form,
\begin{equation}
\small
\begin{aligned}
\mathop{\min}_{u} \lambda {\left\Vert {u}\!-\!{f}\right\Vert }_1\!+\!\alpha  {\left\Vert \nabla u\right\Vert }_1\!+\!\beta{\left\Vert {\nabla}^2 u \right\Vert}_0,
\end{aligned}
\label{Eq:eq5}
\end{equation}
where, $\lambda, \alpha, \beta$ are positive parameters. Notice that $\lambda$ is not necessarily varying in theory, for example, fixing $\lambda=1$, and it is introduced here to rescale Eq. \ref{Eq:eq5} into an appropriate scale for a more accurate numerical solution. Despite the simple form, it enjoys some startling properties for image structure-texture decomposition. We additionally discuss the advantages and benefits of Eq. \ref{Eq:eq5} compared with many existing methods. 

On the one hand, it has been shown that high-quality image decomposition results are not easy to guarantee in many existing methods, including structure-aware filters ~\cite{tomasi1998bilateral, xu2011image}, TV-based regularization~\cite{rudin1992nonlinear} and higher-order extensions~\cite{jung2015simultaneous, bredies2013properties} since they either give over-smoothing results in spikes and sharpening edges --- taking total variation (TV) method~\cite{rudin1992nonlinear} for example, and/or introduce stair-case artifacts in polynomial smoothing surfaces such as the famous $L_0$-norm gradient regularization~\cite{xu2011image}. On the other hand, several textural analysis models~\cite{alliney1997property, meyer2001oscillating, osher2003image} faithfully decouple the large oscillating patterns, while they still suffer from over-smoothing problems, especially around sharpening and discontinuous edges. In a nutshell, many existing methods have limitations in recovering image structures that coexist with sparse features and polynomial-smoothing surfaces, especially in large oscillating textural situations. It is actually difficult for them to weigh a balance to avoid two aspects of deficiencies simultaneously. In contrast, our semi-sparsity model as demonstrated in the below experiments retains edge-preserving properties in strong edges and spikes, and also produces more appropriate results in polynomial-smoothing surfaces, giving a so-called simultaneous-fitting ability in both sharpening and smoothing regions. Moreover, the textural analysis in $L^1$ space also enables us to process image decomposition in large oscillating textures. We conclude that it provides a way to deal with both cases in high-level fidelity, which makes it essentially different from the traditional filtering methods.

\subsection{Differing from Existing Higher-order Methods}

There are two existing works highly related to the proposed semi-sparsity regularization model, in particular, the total generalization variation (TGV)~\cite{bredies2010total} and high-order total variations~\cite{bergounioux2010second, 
papafitsoros2014combined}, as well as many of their extensions~\cite{bredies2013properties, jung2015simultaneous, papafitsoros2015novel}. It is necessary to discuss them and show their differences both theoretically and experimentally.

Mathematically, the TGV-based model~\cite{bredies2010total} suggested the following regularization, 
$$
\operatorname{TGV}(u)=\min_u \alpha_1 {\left\Vert \nabla u-w\right\Vert}_1 +\alpha_2 {\left\Vert \xi(w)\right\Vert}_1,
$$
where $\alpha_1, \alpha_2 \in \mathbb{R}^{+}$are weights, and $\xi(w)=\frac{1}{2}\left(\nabla w+\nabla w^T\right)$ denotes the distributional symmetrized derivative. Such a regularization is always combined with different textural analysis functions~\cite{bredies2013properties, jung2015simultaneous} to reduce the staircase artifacts in many total variational (TV) methods.

The TGV-based method is essentially a second-order regularization model by introducing an auxiliary variable $w$. Notice that $\nabla u \approx w$  when $\alpha_1 \to \infty$, while the exact equality is not attainable in practice, thus there exists a discrepancy between $\nabla u$ and $w$, in particular around sharpening edges. In this case, $w$ can be viewed as a smoothing approximation of $\nabla u$. Notice also that the second term $\alpha_2 {\left\Vert \xi(w)\right\Vert}_1$ also has a smoothing role when imposing a regularization for $w$, which may lead to over-smoothing results in sharpening edges of $u$. The defect may be alleviated by increasing $\alpha_1$ and reducing $\alpha_2$, but it is not easy to find a balance in case of varying scales of oscillating textures. The intuitive conclusion is also observed in our visual comparisons. As illustrated therein, it is usually difficult to find a group of $\alpha$ and $\beta$ to preserve the strong edges precisely.

The $\mathrm{TV}\text{-}\mathrm{TV}^2$ regularization is another higher-order method that has drawn attention in image restoration~\cite{bergounioux2010second, papafitsoros2014combined}, which is also proposed as a higher-order extension of the $\mathrm{TV}$ regularizer to overcome the staircase artifacts. By definition, the $\mathrm{TV}\text{-}\mathrm{TV}^2$ regularization has the form,
$$
\mathrm{TV}\text{-}\mathrm{TV}^2(u)=\min_u \alpha_1 {\left\Vert \nabla u\right\Vert}_1 +\alpha_2 {\left\Vert \nabla^2 u\right\Vert}_1,
$$
The first term $\alpha_1 {\left\Vert \nabla u\right\Vert}_1$ is a $\mathrm{TV}$-based regularization that encourages piece-wise constant $u$, and the second term $\alpha_2 {\left\Vert \nabla^2 u\right\Vert}_1$ plays a similar role but acts on the second-order gradient $\nabla u$. The $\mathrm{TV}\text{-}\mathrm{TV}^2$ regularization helps suppress staircase artifacts since  $\nabla u$ is imposed to be piece-wise constant. However, it is also known that the $\mathrm{TV}$ regularization would cause blur effects around sharpening edges, and the second-order regularization would further strengthen the smoothing effect. In summary, the $\mathrm{TV}\text{-}\mathrm{TV}^2$ regularization indeed helps reduce the stair-case artifacts but the blur results in strong edges are also not avoidable, especially in cases of large oscillating textures or strong noise. Essentially, it has a similar property as the TGV regularization.

The semi-sparsity regularization inherits the advantages of higher-order methods in terms of mitigating stair-case artifacts. In parallel, the inclusion of an $L_0$-norm constraint in higher-order gradients also enables the restoration of sharpening edges without compromising the fitting ability of higher-order models in polynomial smoothing surfaces. The analysis of image textures in $L^1$ space further enhances the capacity to decompose large oscillating patterns. In summary, the proposed method combines the advantages of both TGV and $\mathrm{TV}\text{-}\mathrm{TV}^2$, and manages to overcome their weakness, allowing image decomposition for a broad spectrum of natural images.

\section{Efficient Multi-block ADMM Solver}
\label{sec:solution}

The semi-sparsity image decomposition model in Eq. \ref{Eq:eq5} poses challenges in direct solution due to the non-smooth $L^1$-norm and non-convex nature of $L_0$-norm regularization. It has proved that such a minimization problem can be approximately solved based on several efficient numerical algorithms such as iterative thresholding algorithm~\cite{blumensath2009iterative}, half-quadratic (HQ) splitting technique~\cite{wang2008new}, alternating direction method of multipliers (ADMM)~\cite{davis2017three, ono2017l0}, and so on. 

We here employ a multi-block ADMM algorithm to solve the proposed model because it is applicable for a type of non-convex minimization problem~\cite{boyd2011distributed, davis2017three}. Moreover, it is easy to implement and has a low computational cost even for large-scale problems. For completeness, we briefly introduce the multi-block ADMM algorithm with the $M$ separable objective functions and $N$ constraints, 

\begin{equation}
\begin{aligned}
\mathop{\min}_{\{x_i\}} f_{1}\left(x_{1}\right)+f_{2}\left(x_{2}\right)+\cdots&+f_{M}\left(x_{M}\right)\\
\textit{s.t.} 
\quad A_{1,1} x_{1}+A_{1,2} x_{1,2}+\cdots&+A_{1,M} x_{M}=b_{1}\\
\vdots \; &  \\
\quad A_{N,1} x_{1}+A_{N,2} x_{N,2}+\cdots&+A_{N,M} x_{M}=b_{N},
\end{aligned}
\label{Eq:eq6}
\end{equation}
where $x_{i} \in \mathcal{X}_{i} \subset \mathbb{R}^{n_{i}}$ are closed convex sets, the coefficient matrix $A_{n,i} \in \mathbb{R}^{p \times n_{i}}, b_k \in \mathbb{R}^{p}$ and $f_{i}: \mathbb{R}^{n_{i}} \rightarrow$ $\mathbb{R}^{p}$ are some proper functions.

The multi-block ADMM algorithm is a general case of the ADMM algorithm. Accordingly, we rewrite the Eq. \ref{Eq:eq6} in its augmented Lagrangian form,
\begin{equation}
\begin{aligned}
\mathcal{L}_{\rho} \left(\mathbf{x}, \mathbf{z}\right):=&\sum_{i=1}^{M} f_{i}\left(x_{i}\right)-\sum_{n=1}^{N}\left\langle z_n, \sum_{i=1}^{M} A_{n,i} x_{i}-b_{j}\right\rangle\\+&\sum_{n=1}^{N} 
 \left(\frac{\rho_n}{2}\left\|\sum_{i=1}^{M} A_{n,i} x_{i}-b_{j}\right\|^{2}\right),
\end{aligned}
\label{Eq:eq7}
\end{equation}
where $\mathbf{z} \!=\! \{z_1, \cdots, z_N\}$ is the Lagrange multiplier and $\rho \!=\! \{\rho_1, \rho_2 \cdots, \rho_N; \rho_i\!>\!0\}$ is a penalty parameter. The augmented Lagrangian problem can be solved by iteratively updating the following sub-problems:
\begin{equation}
\left\{\begin{aligned}
x_{1}^{k+1} &:=\operatorname{argmin}_{x_{1}} \mathcal{L}_{\rho}\left(x_{1}, x_{2}^{k}, \cdots, x_{M}^{k} ; z^{k}\right) \\
& \quad \vdots \\
x_{M}^{k+1} &:=\operatorname{argmin}_{x_{M}} \mathcal{L}_{\rho}\left(x_{1}^{k+1}, x_{2}^{k+1}, \cdots, x_{M-1}^{k+1}, x_{M} ; z^{k}\right) \\
z_n^{k+1} &:=z_n^{k}-\rho_n \left(\sum_{n=1}^{N} A_{n,i} x_{i}^{k+1}-b_j\right)\\
\end{aligned}\right.
\label{Eq:eq8}
\end{equation}

Now, it is easy to reformulate the proposed semi-sparsity image decomposition model of Eq. \ref{Eq:eq5} into a three-block ADMM model. Specifically, we consider the discrete case of Eq. \ref{Eq:eq5} and formulate it as a constrained optimization problem by introducing the variables $\boldsymbol{h}, \boldsymbol{g}$ and $\boldsymbol{w}$, 
\begin{equation}
    \begin{aligned}
        \mathop{\min}_{\{\boldsymbol{u,h, g, w}\}} \;  & \lambda {\left\Vert \boldsymbol{u} \! - \!\boldsymbol{f} \right\Vert }_1 + \alpha {\left\Vert \mathbf{D}\boldsymbol{u} \right\Vert }_1 +\beta {\left\Vert \mathbf{L}\boldsymbol{u}\right\Vert }_0,\\
        \textit{s.t.} 
        \quad  & \; \boldsymbol{u} \! - \!\boldsymbol{f} \!=\! \boldsymbol{h}, \quad \mathbf{D}\boldsymbol{u} \!= \!\boldsymbol{g}, \quad \mathbf{L}\boldsymbol{u} \! = \!\boldsymbol{w},
    \end{aligned}
    \label{Eq:eq9}
\end{equation}
where $\mathbf{D}\!=\!\{\mathbf{D}_x; \mathbf{D}_y\}$ and $\mathbf{L}\!=\!\{\mathbf{L}_{xx}; \mathbf{L}_{xy}; \mathbf{L}_{yx};\mathbf{L}_{yy} \}$ are the discrete differential operators in the first-order and second-order cases along the $x$- and $y$- directions, respectively. The augmented Lagrangian form of the three-block ADMM model of Eq. \ref{Eq:eq9} has the form,
\begin{equation}
    \begin{aligned}
        \mathcal{L}_{\rho} \left(\boldsymbol{u,h,g,w}; \mathbf{z}\right):=
        & \lambda {\left\Vert \boldsymbol{h} \right\Vert }_1\!+\!\alpha {\left\Vert \boldsymbol{g} \right\Vert }_1 \!+\!\beta {\left\Vert \boldsymbol{w}\right\Vert }_0\\
        -&\left\langle \boldsymbol{z}_1, \boldsymbol{u} \! - \!\boldsymbol{f} \! - \!\boldsymbol{h} \right\rangle+\frac{\rho_1}{2} \left\|\boldsymbol{u} \!-\!\boldsymbol{f} \!-\!\boldsymbol{h}\right\|^{2}_2\\
        -& \left\langle \boldsymbol{z}_2, \mathbf{D}\boldsymbol{u}\!-\!\boldsymbol{g} \right\rangle+\frac{\rho_2}{2} \left\|\mathbf{D}\boldsymbol{u}\!-\!\boldsymbol{g}\right\|^{2}_2\\
        - &\left\langle \boldsymbol{z}_3, \mathbf{L}\boldsymbol{u}\!-\!\boldsymbol{w} \right\rangle\!+\! \frac{\rho_3}{2} \left\|\mathbf{L}\boldsymbol{u}\!-\!\boldsymbol{w}\right\|^{2}_2,
    \end{aligned}
    \label{Eq:eq10}
\end{equation}
where $\mathbf{z} = \{\boldsymbol{z}_1, \boldsymbol{z}_2, \boldsymbol{z}_3\}$ and $\rho = \{\rho_1, \rho_2, \rho_3\}$ are the dual variables and penalty parameters adapted from Eq. \ref{Eq:eq7}. Clearly, the problem Eq. \ref{Eq:eq10} can be solved based on the iterative procedure presented in Eq. \ref{Eq:eq8}. The benefit of Eq. \ref{Eq:eq9} is that each sub-problem can be efficiently solved even for a large-scale problem. We briefly explain the sub-problems and show their property for efficient solutions. 

\textbf{Sub-problem $\boldsymbol{u}$ :} The augmented Lagrangian objective function Eq. \ref{Eq:eq10} reduces to a quadratic function with respect to $\boldsymbol{u}$ when fixing the other variables, which gives rise to an equivalent minimization problem,
\begin{equation}
    \begin{aligned}
        \mathop{\min}_{\boldsymbol{u}} \quad &{\rho_1{\left\Vert \boldsymbol{u}\!-\!\boldsymbol{f}\!-\!\boldsymbol{h}\!-\!\frac{\boldsymbol{z}_1}{\rho_1}\right\Vert }_2^2}+{\rho_2{\left\Vert \mathbf{D}\boldsymbol{u}\!-\!\boldsymbol{g}\!-\!\frac{\boldsymbol{z}_2}{\rho_2}\right\Vert }_2^2}\\
        &+{\rho_3{\left\Vert \mathbf{L}\boldsymbol{u}\!-\!\boldsymbol{w}\!-\!\frac{\boldsymbol{z}_3}{\rho_3}\right\Vert }_2^2} + c,
    \end{aligned}
\label{Eq:eq11}
\end{equation}
where $c$ is a constant not related to $u$. The optimal solution of \ref{Eq:eq10} is then attained by solving the following linear system:
\begin{equation}
    \begin{aligned}
        (\rho_1\mathbf{I}\!+\!\rho_2 \mathbf{D}^T\mathbf{D} \!+\!\rho_3 \mathbf{L}^T\mathbf{L}) \boldsymbol{u} =&	\rho_1(\boldsymbol{f}\!+\!\boldsymbol{s}) \!+\! \rho_2 \mathbf{D}^T\boldsymbol{g} \!+\!\rho_3 \mathbf{L}^T\boldsymbol{w} \\ &+ \boldsymbol{z}_1 \!+\! \mathbf{D}^T \boldsymbol{z}_2 \!+\! \mathbf{L}^T \boldsymbol{z}_3,
    \end{aligned}
    \label{Eq:eq12}
\end{equation}
where $\mathbf{I}$, $\mathbf{D}^T$ and $\mathbf{L}^T$ are identity matrix and the transpose of $\mathbf{D}$ and $\mathbf{L}$. Due to the symmetric and positive coefficient matrix, Eq. \ref{Eq:eq12} can be directly solved using several linear solvers, for example, the Gauss-Seidel method and preconditioned conjugate gradients (PCG) method. However, these methods may be still time-consuming when the problem has a large number of variables in our cases. Notice that the differential operators $\mathbf{D}$ and $\mathbf{L}$ are linear invariant operators, it is possible to use fast Fourier transforms (FFTs) to solve Eq. \ref{Eq:eq12} for further acceleration. It is possible to reduce the computational cost  significantly, in particular encountering a large-scale problem such as millions of variables. 

\textbf{Sub-problems $\boldsymbol{h}$ and $\boldsymbol{g}$:} By analogy, the sub-problems with respect to $\boldsymbol{h}$ and $\boldsymbol{g}$ have the forms
\begin{equation}
    \begin{aligned}
        \mathop{\min}_{\boldsymbol{h}} {\lambda} {\left\Vert \boldsymbol{h} \right\Vert }_1 -\left\langle \boldsymbol{z}_1, \boldsymbol{u} \! - \!\boldsymbol{f} \! - \!\boldsymbol{h} \right\rangle+\frac{\rho_1}{2} \left\|\boldsymbol{u} \! - \!\boldsymbol{f} \! - \!\boldsymbol{h}\right\|^{2}_2\\
    \end{aligned}
    \label{Eq:eq13}
\end{equation}
\begin{equation}
    \begin{aligned}
        \mathop{\min}_{\boldsymbol{g}} {\alpha} {\left\Vert \boldsymbol{g} \right\Vert }_1 -\left\langle \boldsymbol{z}_2, \mathbf{D}\boldsymbol{u} \! - \!\boldsymbol{g}  \right\rangle+\frac{\rho_2}{2} \left\|\mathbf{D}\boldsymbol{u} \! - \!\boldsymbol{g}\right\|^{2}_2\\
    \end{aligned}
    \label{Eq:eq14}
\end{equation}

It is known that the optimization problem of Eq. \ref{Eq:eq13} and \ref{Eq:eq14} is a typical $L^1$-norm regularization minimization that can be efficiently solved for the separable property. Specifically, we reduce them into a one-dimensional minimization problem and estimate each variable $\boldsymbol{h}_i$ and $\boldsymbol{g}_i$ individually, giving the closed-form solutions,
\begin{equation}
    \begin{aligned}
        \boldsymbol{h}&=\mathcal{S}(\boldsymbol{u}\! - \!\boldsymbol{f}\! + \!\frac{\boldsymbol{z}_1}{\rho_1}, \frac{\lambda}{\rho_1}), \\
        \boldsymbol{g}&=\mathcal{S}(\mathbf{D}\boldsymbol{u}\! + \!\frac{\boldsymbol{z}_2}{\rho_2}, \frac{\alpha}{\rho_2}),
    \end{aligned}
    \label{Eq:eq15}
\end{equation}
with the soft shrinkage operator defined as:
$$
\mathcal{S}(x_i,\tau) = \text{sgn}(x_i)\cdot(|x_i|-\tau)_+
$$
where $\text{sgn}$ is a sign function, and $(a)_+$ is defined as 0 if $a<0$ and $a$ otherwise. 
	
\textbf{Sub-problems $\boldsymbol{w}$:} Similarity, the sub-problem with respect to $\boldsymbol{w}$ has the form,
\begin{equation}
    \begin{aligned}
        \mathop{\min}_{\boldsymbol{w}} {\beta} {\left\Vert \boldsymbol{w} \right\Vert }_0 -\left\langle \boldsymbol{z}_3, \mathbf{L}\boldsymbol{u} \! - \!\boldsymbol{w} \right\rangle+\frac{\rho_3}{2} \left\|\mathbf{L}\boldsymbol{u} \! - \!\boldsymbol{w}\right\|^{2}_2\\
    \end{aligned}
    \label{Eq:eq16}
\end{equation}
which is a $L_0$-norm minimization problem and has the same separable property as $L^1$-norm minimization with each individual variable given by the closed-form solution
\begin{equation}
    \begin{aligned}
        \boldsymbol{w}=\mathcal{H}(\mathbf{L}\boldsymbol{u}\! + \!\frac{\boldsymbol{z}_3}{\rho_3}, \frac{\beta}{\rho_3})
    \end{aligned}
    \label{Eq:eq17}
\end{equation}
where $\mathcal{H}(x, \tau)$ is the hard-shrinkage operator defined as:
$$
\begin{aligned}
    \mathcal{H}(x_i,\tau) = 
    \begin{cases}
        0, & |x_i| < \tau,\\
        x_i, & \text{otherwise}.
    \end{cases}
\end{aligned}
$$

The dual variables $\boldsymbol{z}_1, \boldsymbol{z}_2$, and $\boldsymbol{z}_3$ are updated as follows:
\begin{equation}
	\begin{aligned}
		\boldsymbol{z}_1&=\boldsymbol{z}_1 + \rho_1(\boldsymbol{u} \! - \!\boldsymbol{f} \! - \!\boldsymbol{h}), \\
		\boldsymbol{z}_2&=\boldsymbol{z}_2 + \rho_2(\mathbf{D}\boldsymbol{u}\!-\!\boldsymbol{g}), \\
		\boldsymbol{z}_3&=\boldsymbol{z}_3 + \rho_3(\mathbf{L}\boldsymbol{u}\!-\!\boldsymbol{w}). 
	\end{aligned}
	\label{Eq:eq18}
\end{equation}

The three-block ADMM algorithm alternatively solves the sub-problems and updates the Lagrange multipliers until the given stop criteria are met, which leads to an iterative procedure for the proposed image decomposition model. Since all sub-problems have closed-form solutions in low computational complexity, the challenging non-convex $L_0$-norm regularized problem is empirically solvable even under a large-scale of variables. The numerical results will further demonstrate the effectiveness and efficiency of the multi-block ADMM algorithm. 

The  convergent analysis of multi-block ADMM solver can be directly built from the general ADMM case~\cite{ boyd2011distributed}. Notice that Eq. \ref{Eq:eq4} and Eq. \ref{Eq:eq5} have the non-convex $L_0$-norm regularization, it is not easy to construct a convergence analysis. Recently, it has been shown in the study~\cite{wang2019global} that the ADMM algorithm is also applicable for a wide range of non-smooth and non-convex problems. For example, the convergent analysis can be built when one or more of $f_i$ in Eq. \ref{Eq:eq6} is lower semi-continuous with the remainders are the Lipschitz differentiable functions, which is our case and claims the convergence of  \textbf{Algorithm} \ref{Alg:alg1}, as the first and second terms in Eq. \ref{Eq:eq9} are convex $L^1$-norm with Lipschitz continuous property, and the non-convex $L_0$ norm  with has lower semi-continuous property. Actually, it is also easy to see that each sub-problem convergent to its criteria point, which makes it possible to find a Cauchy sequence of the solution $\{u^k\}_{k=0}^{\infty}$. Here, the sub-problem of Eq. \ref{Eq:eq16} is $L_0$-regularization and the iterative hard-thresholding operator is convergence under the Restricted Isometry Property (RIP) condition~\cite{blumensath2009iterative}. The reader may refer to~\cite{davis2017three, wang2019global, han2022survey} for more convergence details.  
\begin{algorithm}[!t]
    \caption{ADMM Solver for Semi-sparse Model}
    \label{Alg:alg1}
    \begin{algorithmic}[1]
        \STATE \textbf{Input:} signal/image $\boldsymbol{f}$, parameters ${\lambda, \alpha, \beta}$, and weights $\rho_1$, $\rho_2$, $\rho_3$;
        \STATE \textbf{Initialization:} $\boldsymbol{u}^0\!\leftarrow\!\boldsymbol{0}$; $\boldsymbol{h}^0, \boldsymbol{g}^0, \boldsymbol{w}^0\!\leftarrow\!\boldsymbol{0}$;  $\boldsymbol{z}^{0}_{1}$, $\boldsymbol{z}^{0}_{2}$, $\boldsymbol{z}^{0}_{3}\!\leftarrow\!0$, and ${k\!\leftarrow\!0}$;
        \WHILE {not ${\left\|u^{k+1}\!-\!u^k\right\|_2^2}/{\left\|u^{\star}\right\|_2^2} \leq \varepsilon$}
        \STATE Solve Eq. \ref{Eq:eq12} for $\boldsymbol{u}^{k+1}$ with the FFT acceleration;
        \STATE Solve Eq. \ref{Eq:eq13} for $\boldsymbol{h}^{k+1}$ with soft-shrinkage operator;
        \STATE Solve Eq. \ref{Eq:eq14} for $\boldsymbol{g}^{k+1}$ with soft-shrinkage operator;
        \STATE Solve Eq. \ref{Eq:eq16} for $\boldsymbol{w}^{k+1}$ with hard-shrinkage operator;
        \STATE Update Eq. \ref{Eq:eq18} for Lagrange multipliers $\boldsymbol{z}^{k+1}_{i}$:\\
        \quad $\boldsymbol{z}^{k+1}_{1}=\boldsymbol{z}^{k}_{1} + \rho_1(\boldsymbol{u}^{k} \! - \!\boldsymbol{f} \! - \!\boldsymbol{h}^{k})$,\\
	\quad $\boldsymbol{z}^{k+1}_{2}=\boldsymbol{z}^{k}_{2} + \rho_2(\mathbf{D}\boldsymbol{u}^{k}\!-\!\boldsymbol{g}^{k})$,\\
	\quad $\boldsymbol{z}^{k+1}_{3}=\boldsymbol{z}^{k}_{3} + \rho_3(\mathbf{L}\boldsymbol{u}^{k}\!-\!\boldsymbol{w}^{k})$;
\STATE Increment $k: k= k+1$,
\ENDWHILE
\STATE {\textbf{Output:}} the structural part ${\boldsymbol{u}}$.
    \end{algorithmic}
\end{algorithm}

\section{Experimental Results}
\label{sec:experimental_results}

In this section, we further demonstrate our semi-sparse model and its high-quality decomposition performance in scenarios of images composed of different structures and textures. We start the discussion with the interpretations of a new dataset, parameter configurations, compared methods, and so on. A series of numerical results are also presented and compared with the state-of-the-art methods both qualitatively and quantitatively. 

\begin{figure}[!t]
	\begin{center}
		{\includegraphics[width=0.5\textwidth]{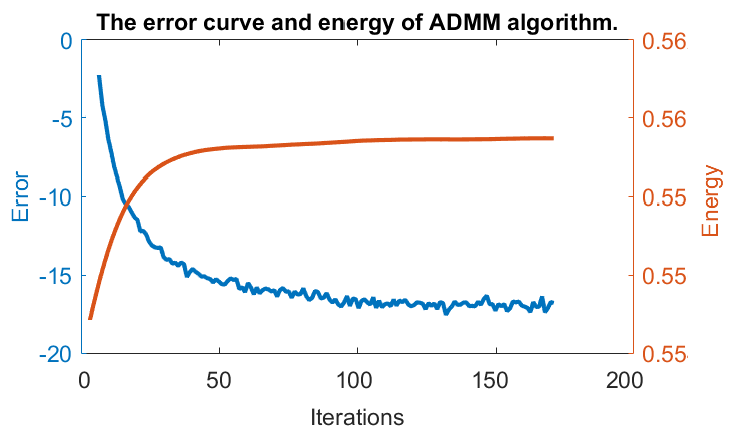}}
	\end{center}
	\caption{The (log) \textbf{error curve} and \textbf{energy} of $u^k$ during the ADMM iterative procedure.}
    \label{Fig:fig5}
\end{figure}

\subsection{Dataset and Compared Methods}

For the sake of comprehensive analysis, we first collect a new dataset for image structure and texture decomposition. This dataset has 32 images chosen from the community of image decomposition~\cite{xu2012structure, cho2014bilateral, kodak}.  As shown in the later experiments, the dataset consists of a set of images that contain various structural patterns such as sharpening edges, piece-wise constant and smoothing surfaces, as well as diverse textural patterns with large oscillating values, multiple scales and directions, and so on. The inclusion of such images allows us to thoroughly assess the performance of our method across different types of visual content.

To ensure a systematic comparison, we elaborate on our semi-sparsity model with its superior performance against a series of image decomposition methods. Specifically, we compare the model with the well-known $\mathrm{ROF}$ model~\cite{rudin1992nonlinear}, $\mathrm{TV}$-$L^1$~\cite{le2014cartoon+}, $\mathrm{TV}$-$G$~\cite{meyer2001oscillating}, $\mathrm{TV}$-$\mathcal{H}$~\cite{osher2003image}, $\mathrm{TV}$-$G$-$\mathcal{H}$~\cite{xu2022new}, BTF~\cite{sun2017image}, RTV~\cite{xu2012structure}, $\mathrm{TGV}$-$L^1$~\cite{bredies2013properties}, $\mathrm{TGV}$-$\mathcal{H}$~\cite{jung2015simultaneous}, $\mathrm{HTV}$-$\mathcal{H}$~\cite{jung2015simultaneous}, which cover most of the structural and textual models listed in Tab. \ref{Tab:tab1} and Tab. \ref{Tab:tab2}. As illustrated hereafter, we will consider three typical scenarios consisting of different image structures and textures: 1) piece-wise constant structures with uniform but large oscillating textures, 2) piece-wise smoothing structures with multi-scale oscillating textures, and 3) a mixed case with complex structural and textural patterns.

\subsection{Parameters and Configurations}

Recalling the model in Eq. \ref{Eq:eq4} and \ref{Eq:eq5}, $\lambda$, $\alpha$, and $\beta$ control the amount of smoothness penalty applied to output image structures. As aforementioned, $\lambda$ is introduced to rescale Eq. \ref{Eq:eq5} into an appropriate scale, depending on the precision of discrete images and the threshold operators in Eq. \ref{Eq:eq15} and \ref{Eq:eq17}. For 8-bit color images\footnote{All images are normalized into the range $[0, 1]$ in our experiments.}, we empirically found that $\lambda \!\in\! [0.0001, 1.0]$, $\alpha \!\in\! [0.0001, 0.1]$ and $\beta \!\in\! [0.0001, 0.1]$ are always suitable for a high-accuracy numerical solution when taking into account weights $\rho_i \!=\!1 (i\!=\!1,2,3)$ without any specification. The parameters may vary from image to image depending on the density and oscillating level of image textures. The above configurations provide a good trade-off between accuracy and computational efficiency.

\begin{figure*}[!t]
	\begin{center}
 	\subfloat[Input]
		{\includegraphics[width = 0.24\textwidth]{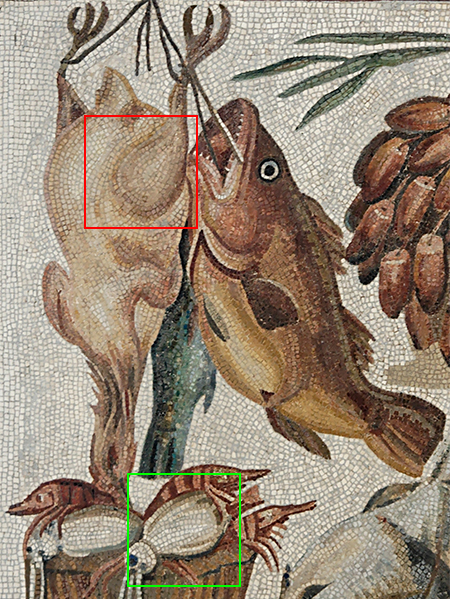}}\hfil
        \subfloat[$\alpha\!=\!0.01, \beta\!=\!0.01$]
		{\includegraphics[width = 0.24\textwidth]{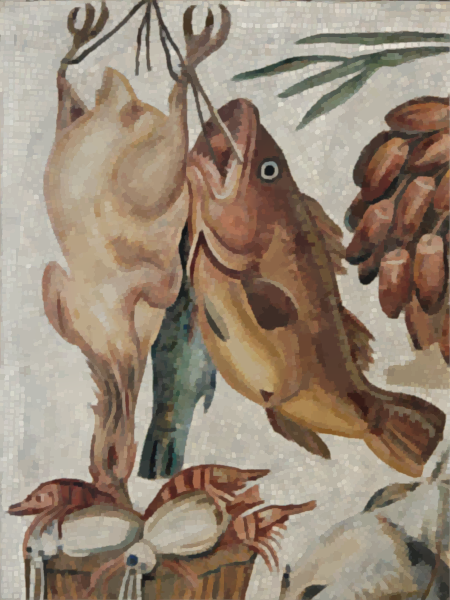}}\hfil
		\subfloat[$\alpha\!=\!0.02, \beta\!=\!0.01$]
		{\includegraphics[width = 0.24\textwidth]{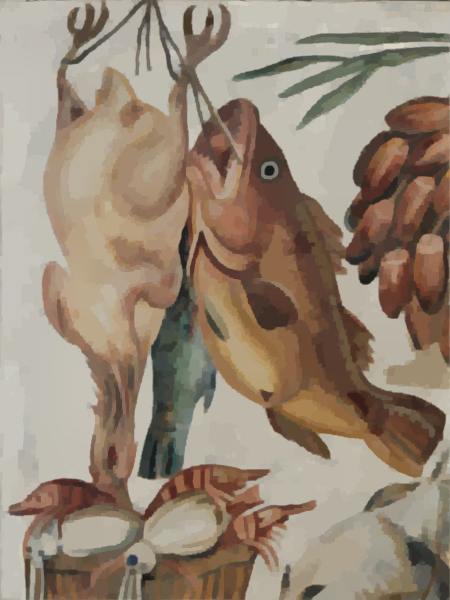}}\hfil
		\subfloat[$\alpha\!=\!0.03, \beta\!=\!0.01$]
		{\includegraphics[width = 0.24\textwidth]{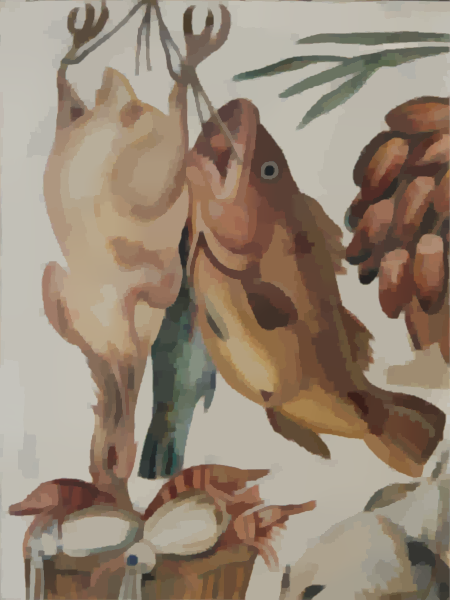}}\hfil
		\subfloat[Close-ups]
		{\includegraphics[width = 0.24\textwidth]{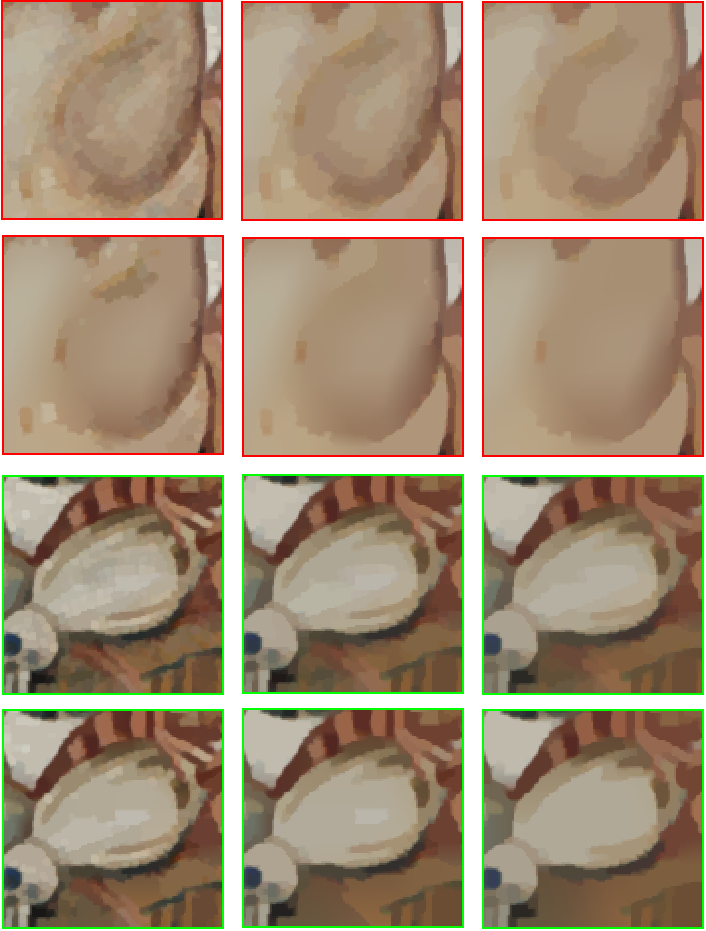}}\hfil
		\subfloat[$\alpha\!=\!0.01, \beta\!=\!0.02$]
		{\includegraphics[width = 0.24\textwidth]{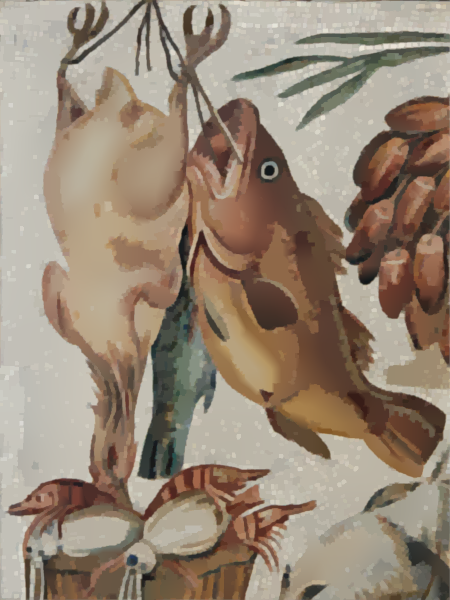}}\hfil
		\subfloat[$\alpha\!=\!0.02, \beta\!=\!0.02$]
		{\includegraphics[width = 0.24\textwidth]{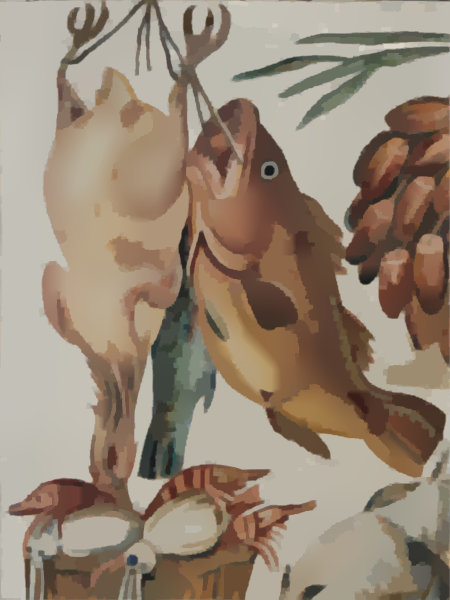}}\hfil
		\subfloat[$\alpha\!=\!0.03, \beta\!=\!0.02$]
		{\includegraphics[width = 0.24\textwidth]{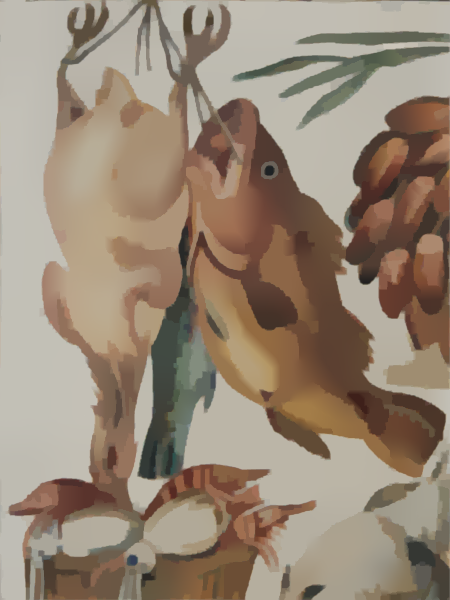}}\hfil
	\end{center}
	\caption{Visual comparison of image structures  with different parameters. $\lambda =0.01$ is fixed here and $\alpha$ is gradually increased in each row to give more and more smoothing results, while a similar smoothing trend is also observed with increasing $\beta$ in each column. Notice that $\alpha$ mainly controls the global smoothness of output structures, and $\beta$ tends to smooth the local features such as staircase results in polynomial smoothing surfaces.} 
 \label{Fig:fig6}
\end{figure*}

For clarity, we show the role of ${\lambda, \alpha}$ and ${\beta}$ in determining the characteristics of output image structures. As illustrated in Fig. \ref{Fig:fig6}, we fix $\lambda \!=\!0.01$ and compare the output structures with varying $\alpha$ and $\beta$. The output image structures in each row tend to be more and more smoothing when increasing $\alpha$, but the stair-case artifacts can not be avoided effectively in some poly-nominal smoothing regions, as $\alpha$ mainly controls the first-order gradient to be piece-wise constant. In contrast, the staircase artifacts can be alleviated when increasing $\beta$ in each column. This conclusion also provides the guidance for choosing $\lambda$, $\alpha$, and $\beta$ in practice --- that is, $\alpha$ is first adjusted to produce acceptable decomposition results and then fine-tuning $\beta$ to reduce the remaining stair-case artifacts to reach better local results.

To demonstrate the efficiency of ADMM iterative procedure, we show its convergence based on the relative error $Q_r^k ={\left\|u^{k+1}\!-\!u^{k} \right\|_2^2}/{\left\|u^\star\right\|_2^2}$. Notice that an ``exact'' solution $u^\star$ is not available here, we instead take an extremely tight stop tolerance, for example, $\varepsilon= 10^{-16}$, and treat the output as $u^\star$ for evaluation. Afterward, the algorithm is rerun under the same configuration with a smaller stop criterion $\varepsilon\!=\!1.0 \times 10^{-12}$ in \textbf{Algorithm} \ref{Alg:alg1}. We also take into account the energy $E_{u^k} = \left\|u^k\right\|_2^2$ to show the decomposition performance. The error curve $Q_r$ and energy $E_{u}$ are plotted in Fig. \ref{Fig:fig5}. As we can see, both of them converge quickly to the stable solution and it roughly needs 20 $\sim$ 50 iterations to produce high-quality image decomposition results with appropriate parameters.

Benefiting from the multi-block ADMM procedure, the proposed model can be solved efficiently. As shown in Algorithm~\ref{Alg:alg1}, the process has three main phases: linear system solver, soft/hard-threshold shrinking operators, and the update of Lagrangian multiplier in each iteration. The computation is dominated by the linear system solver, while it can be also efficiently computed with the FFT acceleration because both the FFT and its inverse transform have the computational cost of $O(N log(N))$, where $N$ is the  number of pixels of an image. The threshold shrinking operator can be computed independently for their separable property, and the Lagrange multiplier can be also updated in-place efficiently. As a result, the number of iterations controls the total time of the proposed method. The quantitative analysis of computational cost is also compared with the other methods in the next section.

\begin{figure*}[!t]
    \begin{center}
        \subfloat[Input]
        {\includegraphics[width=0.24\textwidth]{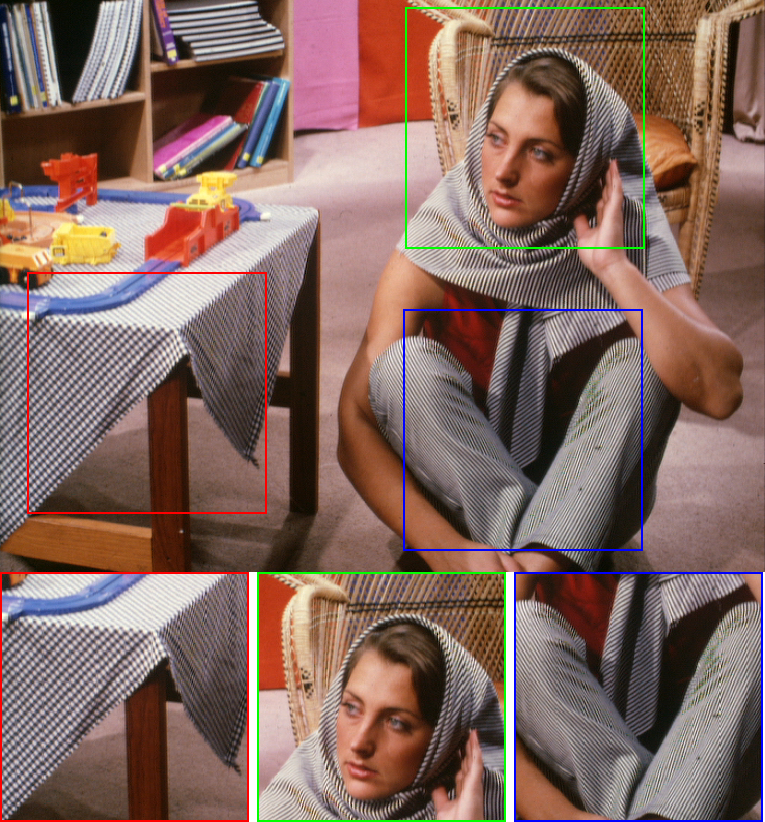}}\hfil
        \subfloat[ROF~\cite{rudin1992nonlinear}]
        {\includegraphics[width=0.24\textwidth]{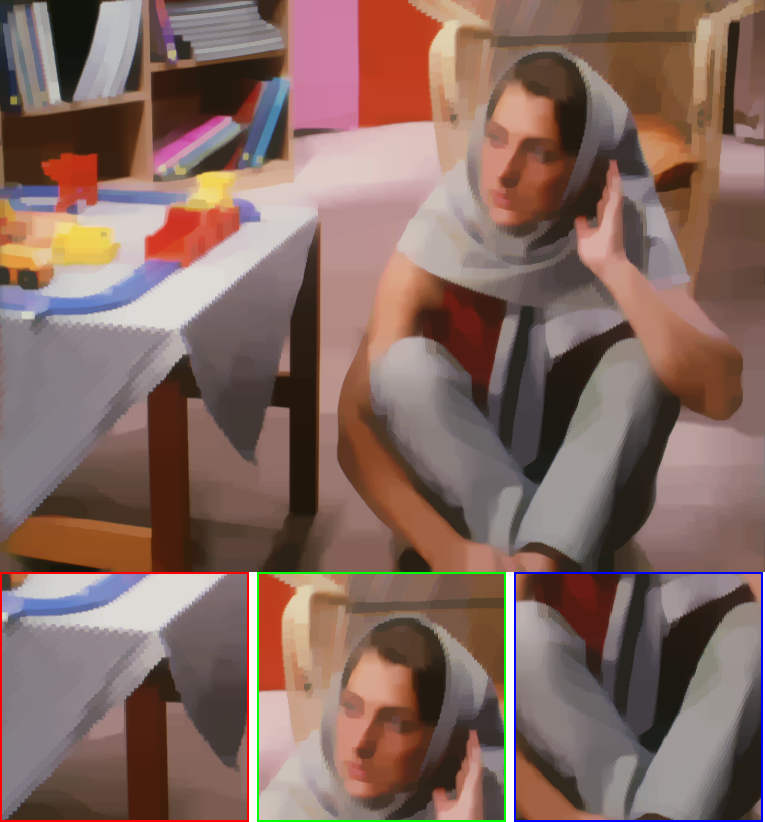}}\hfil
        \subfloat[$\mathrm{TV}$-$L^1$~\cite{le2014cartoon+}]
        {\includegraphics[width=0.24\textwidth]{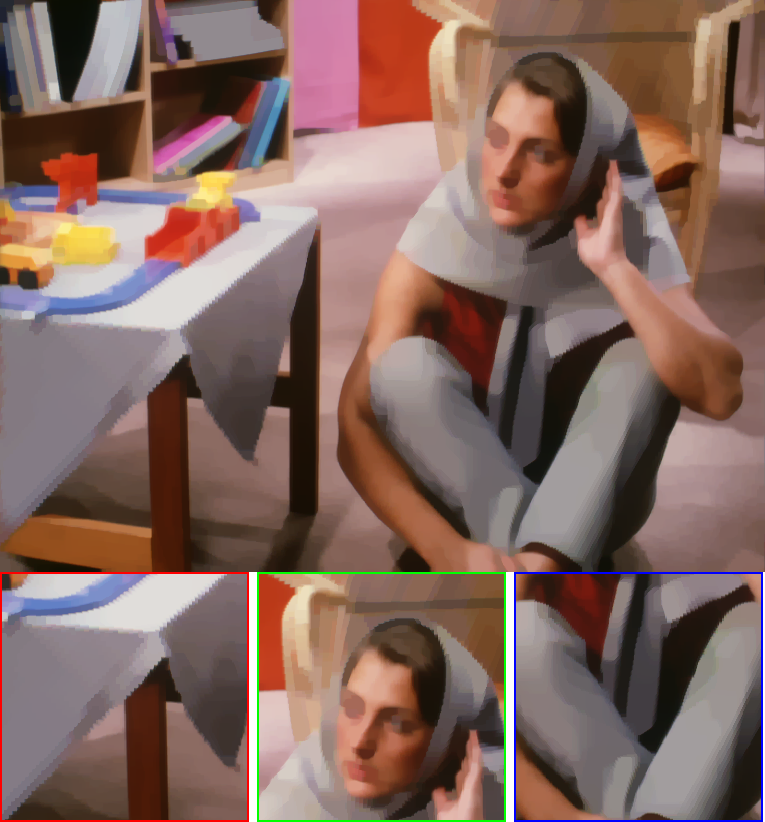}}\hfil
        \subfloat[$\mathrm{TV}$-$G$~\cite{meyer2001oscillating}]
        {\includegraphics[width=0.24\textwidth]{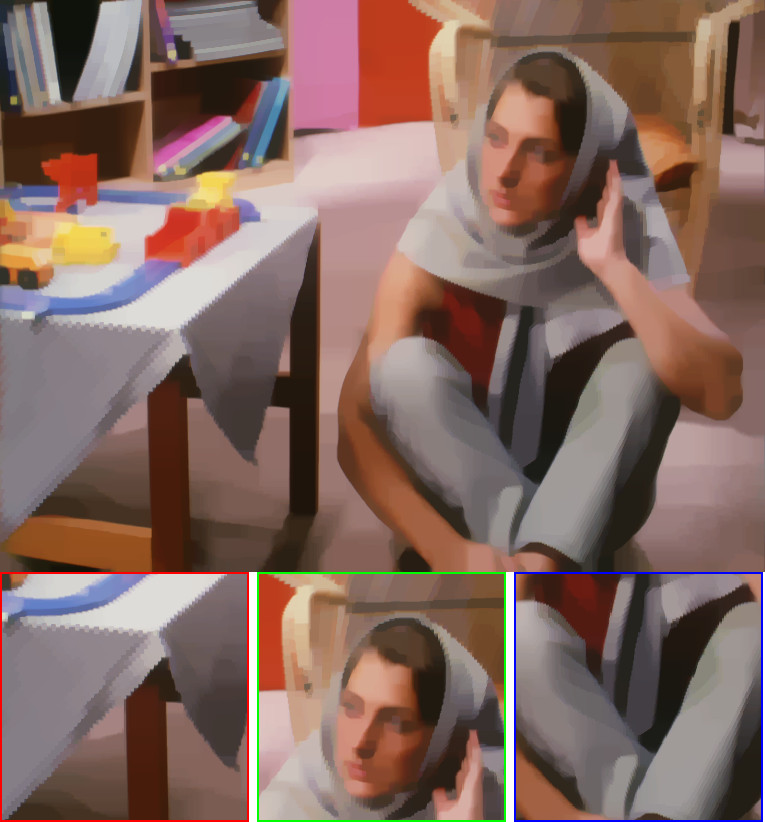}}\hfil
        \subfloat[$\mathrm{TV}$-$\mathcal{H}$~\cite{osher2003image}]
        {\includegraphics[width=0.24\textwidth]{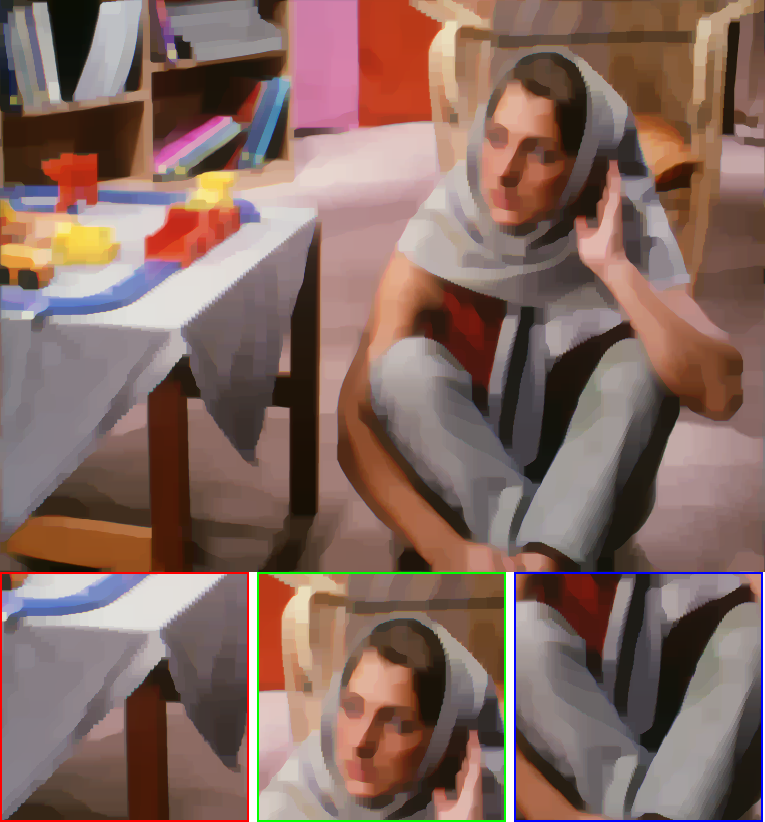}}\hfil
        \subfloat[$\mathrm{TV}$-$G$-$\mathcal{H}$~\cite{xu2022new}]
        {\includegraphics[width=0.24\textwidth]{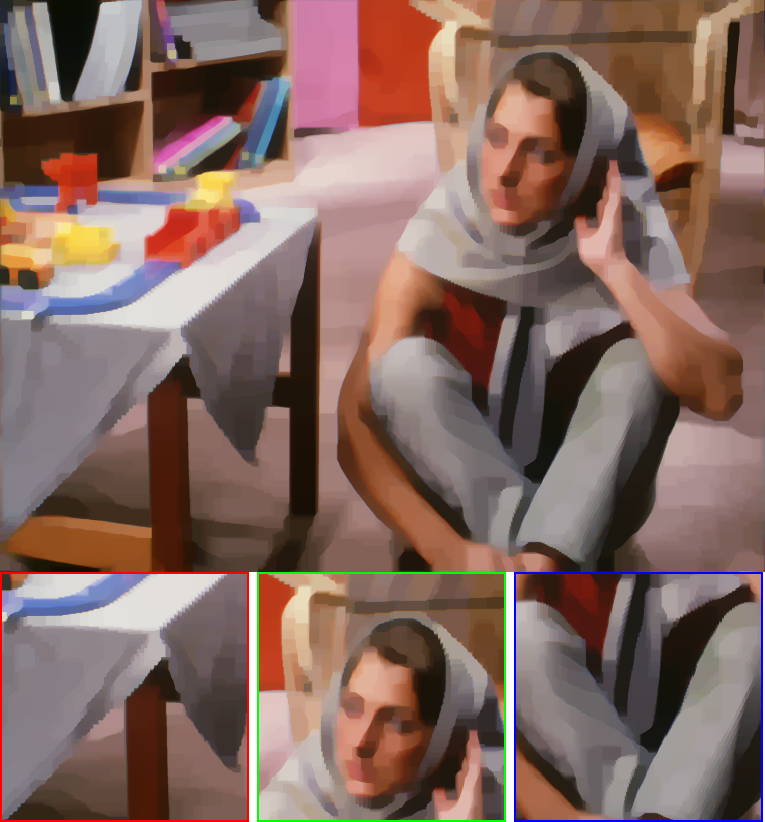}}\hfil
        \subfloat[BTF~\cite{cho2014bilateral}]
        {\includegraphics[width=0.24\textwidth]{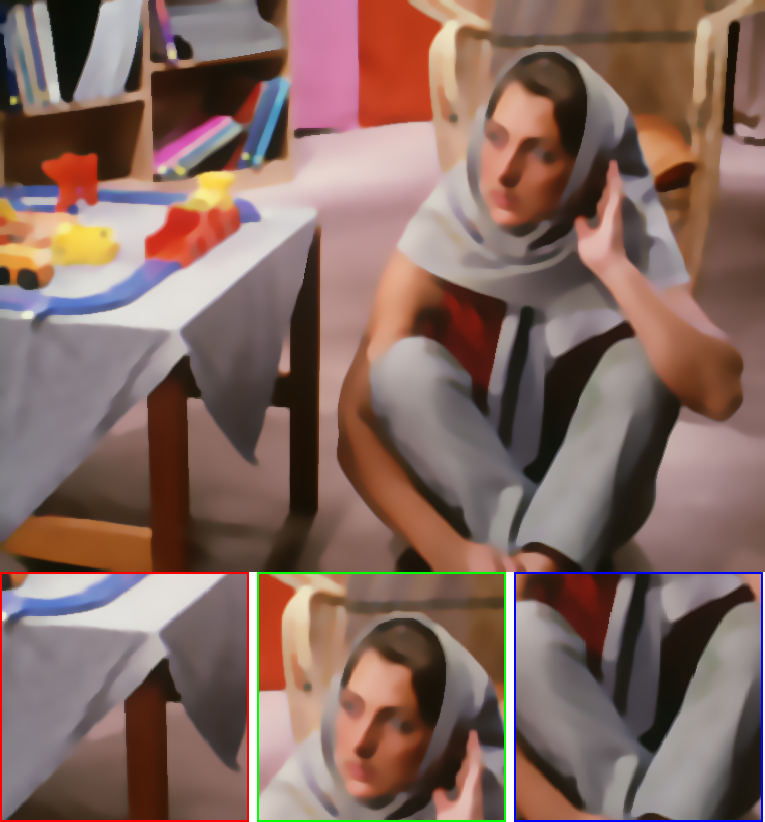}}\hfil
        \subfloat[RTV~\cite{xu2012structure}]
        {\includegraphics[width=0.24\textwidth]{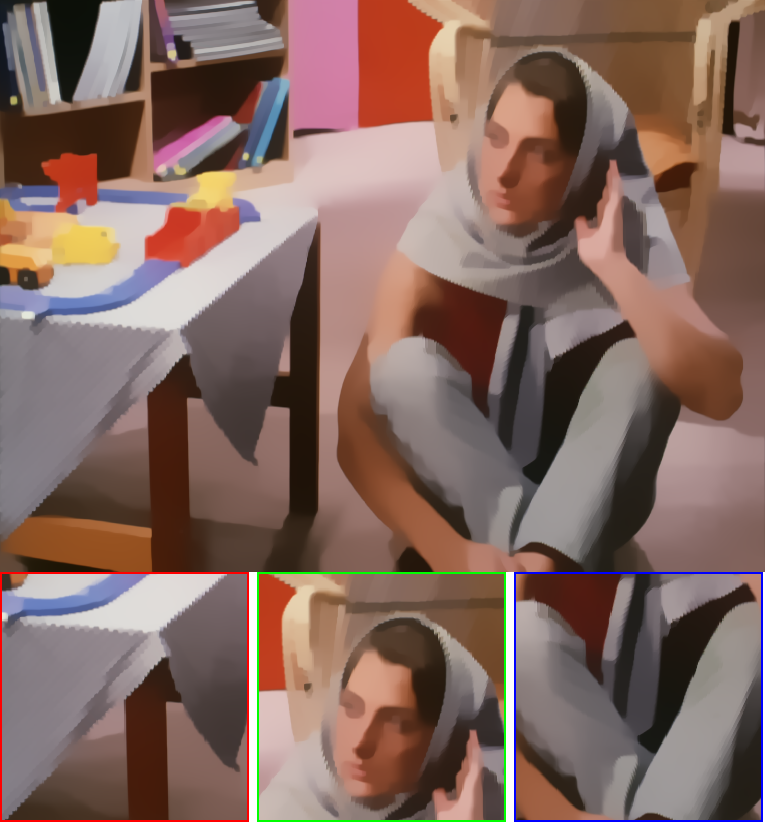}}\hfil
        \subfloat[$\mathrm{TGV}$-$L^1$~\cite{bredies2013properties}]
        {\includegraphics[width=0.24\textwidth]{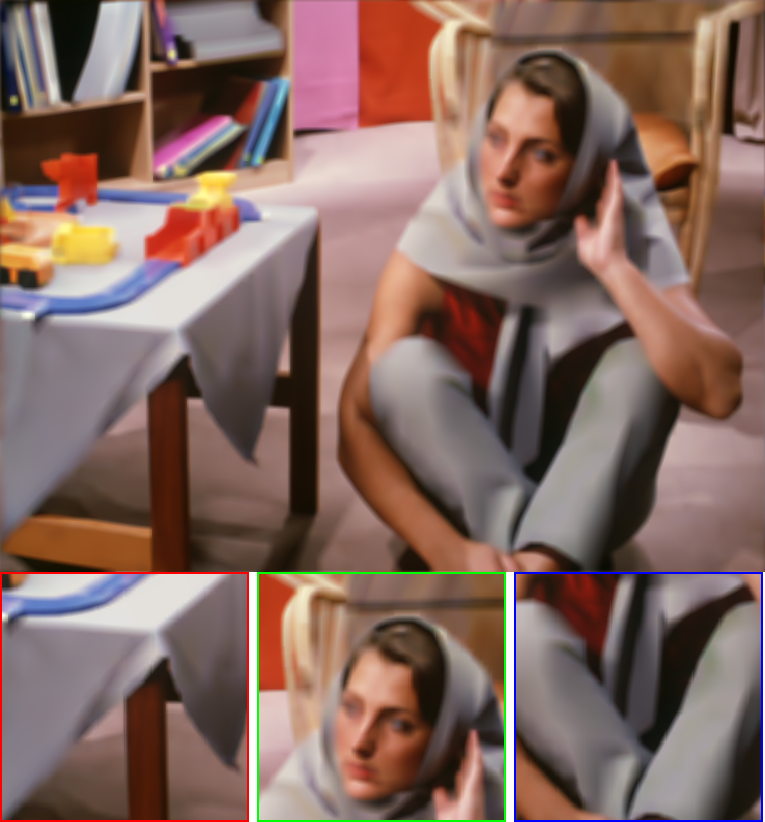}}\hfil
        \subfloat[$\mathrm{TGV}$-$\mathcal{H}$~\cite{jung2015simultaneous}]
        {\includegraphics[width=0.24\textwidth]{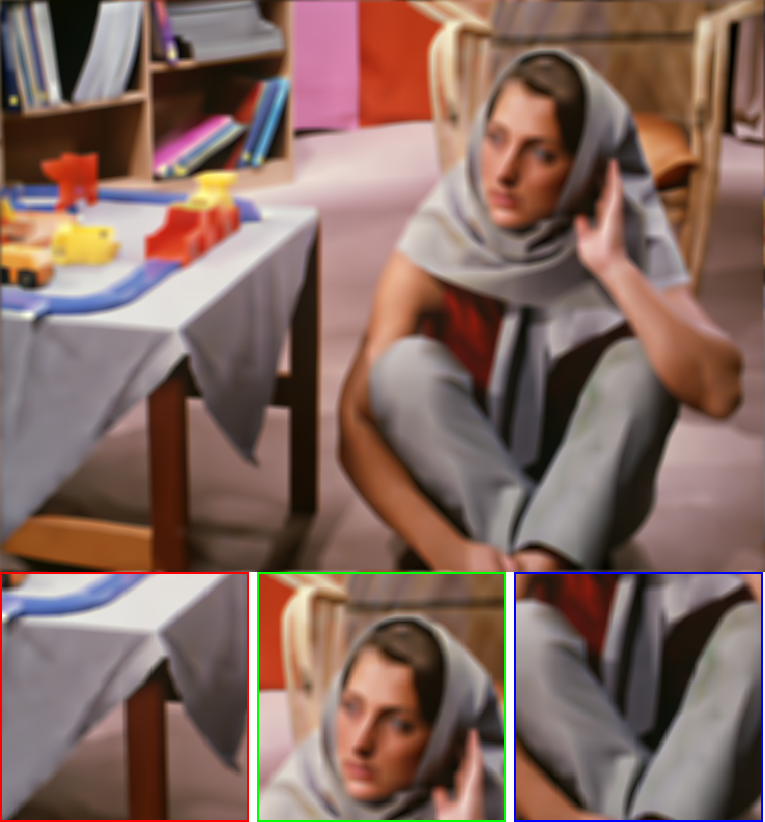}}\hfil
        \subfloat[$\mathrm{HTV}$-$\mathcal{H}$~\cite{jung2015simultaneous}]
        {\includegraphics[width=0.24\textwidth]{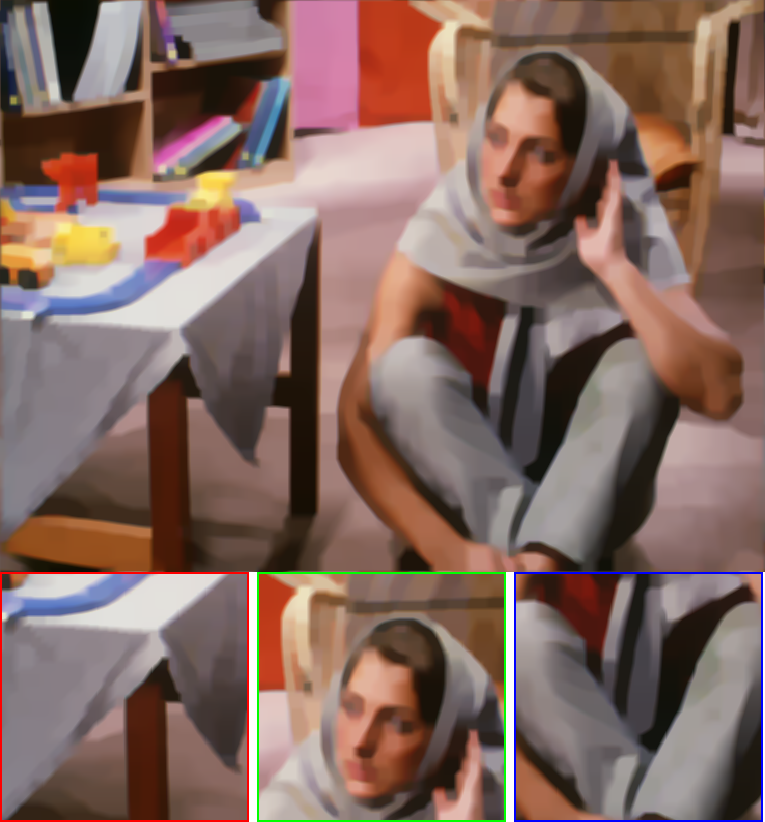}}\hfil
        \subfloat[Ours]
        {\includegraphics[width=0.24\textwidth]{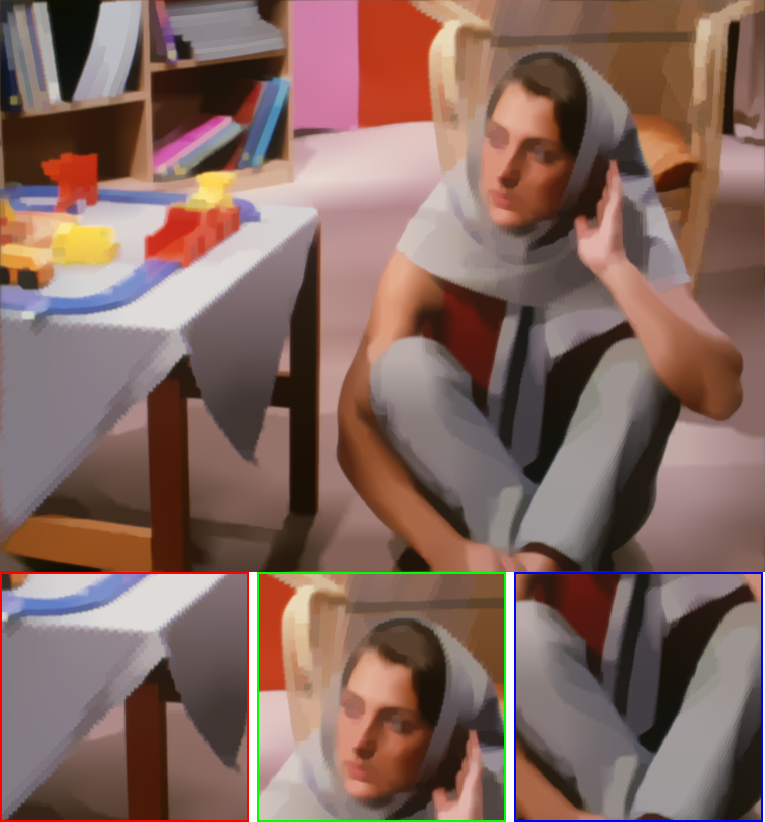}}\hfil
    \end{center}
    \caption{Visual results of image structures in case of \textbf{hybrid and complicated textural and structural components}. 
    (a) Input,  (b) ROF ($\lambda\!=\!0.0045, \alpha\!=\!0.001$)~\cite{rudin1992nonlinear},  (c) $\mathrm{TV}$-$L^1$ ($\lambda\!=\!0.001, \alpha\!=\!0.002$)~\cite{le2014cartoon+},  (d) $\mathrm{TV}$-$G$ ($\lambda\!=\!0.003, \alpha\!=\!0.0006, \gamma\!=\!0.0004$)~\cite{meyer2001oscillating},  (e) $\mathrm{TV}$-$\mathcal{H}$ ($\lambda\!=\!0.001$, $\alpha\!=\!0.01$)~\cite{osher2003image},  (f) $\mathrm{TV}$-$G$-$\mathcal{H}$ ($\lambda\!=\!0.004, \alpha\!=\!0.008, \gamma\!=\!0.005$)~\cite{xu2022new},  (g) BTF ($\sigma\!=\!7.0, iter\!=\!3$)~\cite{cho2014bilateral},  (h) RTV ($\lambda\!=\!0.012, \sigma\!=\!3.0$)~\cite{xu2012structure},  (i) $\mathrm{TGV}$-$L^1$ ($\lambda\!=\!0.001, \alpha\!=\!0.002, \beta\!=\!0.002$)~\cite{bredies2013properties}, (j) $\mathrm{TGV}$-$\mathcal{H}$ ($\lambda\!=\!0.002, \alpha\!=\!0.045, \beta\!=\!0.025$)~\cite{jung2015simultaneous},  (k) $\mathrm{HTV}$-$\mathcal{H}$ ($\lambda\!=\!0.004$, $\alpha\!=\!0.006, \beta\!=\!0.0015$)~\cite{jung2015simultaneous}, and (l) Ours ($\lambda\!=\!0.002, \alpha\!=\!0.005, \beta\!=\!0.001$). Quantitative results with the $STR (C_0/C_1)$ metrics, (b)$\scriptsize{\sim}$(j): 22.84 (0.1908/0.0918),	22.83 (0.0550/0.1161),	22.89 (0.1771/0.0954),	22.78 (0.0392/0.0424),	22.81 (0.0416/0.0326), 22.88 (0.0271/0.0526),	22.85 (0.1371/0.0447),	22.84 (0.0077/0.1087),	22.78 (0.0370/0.1994),	22.85 (0.0507/0.0869),	23.01 (0.0774/0.0644). (Zoom in for better view.)} 
    \label{Fig:fig7}
\end{figure*} %

\subsection {Structure Extraction Performance}

To demonstrate the benefits of the proposed method, we conduct a comparative analysis of the decomposed results. Firstly, we introduce image structure-to-texture ratio (STR) as an objective index to evaluate the smoothness of output structures, which is defined in decibels (dB) as:
$$
STR = 10 \log_{10}\frac{\left\|u \right\|_2^2}{\left\|v\right\|_2^2},
$$
where $u$ and $v$ represent the decomposed image structures and textures, respectively. The STR index has the same definition as signal-to-noise (SNR) in our case when treating $v=f-u$ as noise. For fairness, all compared methods are either configured with a greedy strategy to produce visual-friendly results or fine-tuned by hand to reach a similar smoothing level of image structures evaluated by the STR index. In what follows, we show the structural decomposition results and compare them in different scenarios of image structures and textures.

\begin{figure*}[!t]
	\begin{center}
		{\includegraphics[width=0.19\textwidth]{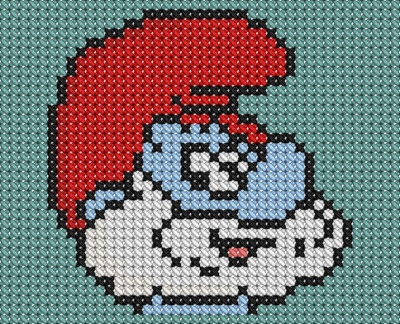}}\hfil
		{\includegraphics[width=0.19\textwidth]{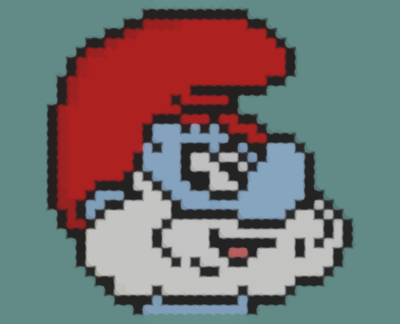}}\hfil
		{\includegraphics[width=0.19\textwidth]{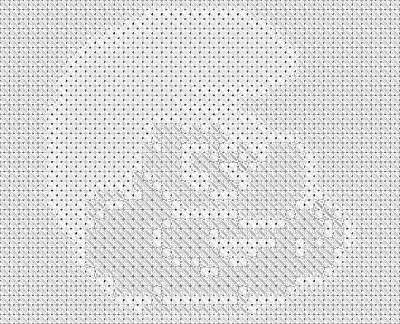}}\hfil
		{\includegraphics[width=0.19\textwidth]{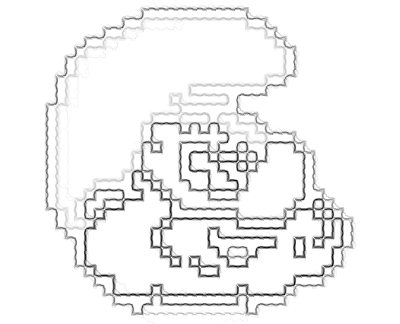}}\hfil
		{\includegraphics[width=0.19\textwidth]{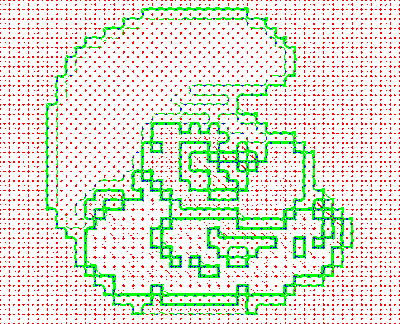}}\hfil
            \subfloat[Input $f$]
		{\includegraphics[width=0.19\textwidth]{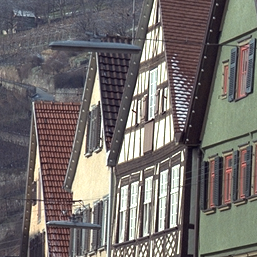}}\hfil
		\subfloat[Structure $u$]
		{\includegraphics[width=0.19\textwidth]{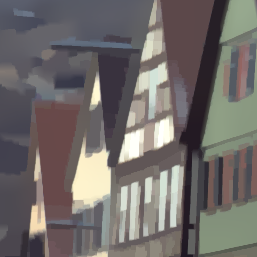}}\hfil
		\subfloat[Texture $v$]
		{\includegraphics[width=0.19\textwidth]{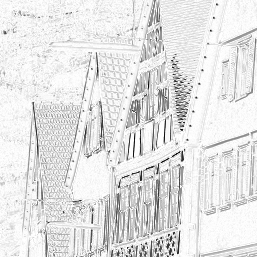}}\hfil
		\subfloat[Gradient $\nabla u$]
		{\includegraphics[width=0.19\textwidth]{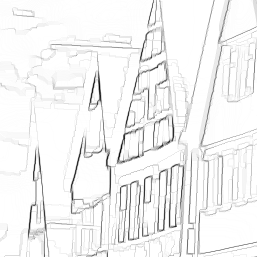}}\hfil
		\subfloat[Color-map]
		{\includegraphics[width=0.19\textwidth]{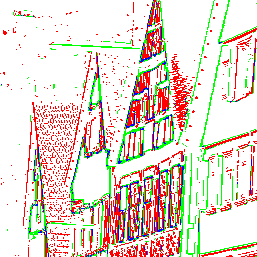}}\hfil
	\end{center}
	\caption{Visual illustrations of the correlation between image structures $u$ (gradient $\nabla u$) and textures $v$. Given (a) input images $f$, our method produce (b) structures $u$ and (c) textures $v=f-u$ and (d) gradients $\nabla u$, which are also plotted in (e) in \textcolor{red}{Red} and \textcolor{green}{Green} to indicate their tiny overlap regions in \textcolor{blue}{Blue}. The correlation coefficients $C_0 = 0.001$ and $C_1= 0.001$. (c)$\sim$(e) are scaled for better view.}
    \label{Fig:fig8}
\end{figure*} %

In Fig. \ref{Fig:fig3}, we show the decomposition structures for the first type of image. In this situation,  it is usually difficult for traditional structure-aware filtering methods\cite{tomasi1998bilateral, xu2011image} to give high-quality results, because of the presence of large oscillating textures. The TV-based models such as TV-$G$~\cite{meyer2001oscillating}, TV-L1~\cite{chambolle2011first} and TV-$\mathcal{H}$~\cite{osher2003image}, in contrast, demonstrate their reasonable performance in achieving fine-balanced decomposition results. The benefits mainly arise from the capacity of TV space in capturing piece-wise constant structures in spite of large oscillating textures. The special-designed BTF method\cite{cho2014bilateral} and RTV model~\cite{xu2012structure} are known for their ability to persevere strong edges when removing large oscillating textures. Despite much better performance, they may introduce slightly blurry effects around strong edges, in particular, the BTF method for the nature of using local spatial weights to suppress periodic textures. Notice also in Fig. \ref{Fig:fig3} that the RTV algorithm may encounter difficulties when removing (white) spikes next to strong edges. The higher-order TGV-$\mathcal{H}$ and HTV-$\mathcal{H}$ models~\cite{jung2015simultaneous} exhibit effective removal of large oscillating textures but suffer from noticeable degradation in sharpening edges. This partly arises from the fact that the TGV regularization tends to penalize the second-order gradients more than the first-order case, leading to a loss of edge fidelity. In contrast, our method receives comparable results in removing large-scale oscillating textures with the best performance in views of edge fidelity.

We continuously analyze the second type of images by considering piece-wise constant and smoothing surfaces in structures, and multi-scale oscillating patterns in textures. As shown in Fig. \ref{Fig:fig4}, the TV-based ROF model~\cite{rudin1992nonlinear}, TV-$G$~\cite{meyer2001oscillating} and TV-L1~\cite{chambolle2011first} show the ability to decouple the large oscillating textures in piece-wise constant backgrounds but they suffer from different levels staircase effects in the polynomial-smoothing (face) surfaces. In this scenario, the BTF method gives a similar result as the TV-based methods with weak stair-case artifacts, but the over-blurring effect still exists in sharpening edges. The RTV model~\cite{xu2012structure}  performs much better results than other existing methods, achieving minimal staircase effects in the face regions while preserving strong edges effectively. The higher-order TGV-$\mathcal{H}$~\cite{jung2015simultaneous} and HTV-$\mathcal{H}$~\cite{jung2015simultaneous} partially suppress the staircase effects but give rise to serious blur degradation around strong edges, which is similar to the case in Fig. \ref{Fig:fig3}. In contrast, our semi-sparsity method demonstrates significantly improved performance as the  RTV model, receiving the desired smoothing results in the face regions and comparable sharpening edges.

The advantage of our semi-sparsity method is further demonstrated with a more challenging case in Fig. \ref{Fig:fig7}. The test image here exhibits highly complex structural and textural patterns, in particular, the oscillating textures are anisotropic in directions, and have coarse-to-fine scales, and varying amplifications, which poses a significant challenge for many image decomposition models. In this case, existing methods either reveal over-smoothing results in texture-less regions or can not remove large oscillating textures, especially in view of the sharpness of discontinuous boundaries among the piece-wise constant/smoothing surfaces. Instead, the proposed semi-sparsity decomposition method gives more preferable results in this complex situation with more clean and smooth structural backgrounds and precise sharpening edges. In summary, the proposed method is applicable to a wide range of natural images and can successfully preserve sharp edges while effectively eliminating the undesired oscillating textures. The advance is mainly beneficial from the leveraging of higher-order $L_0$ regularization for image structures and the measurement of image textures in $L^1$ space.

\subsection{Quantitative Evaluation} 

We further take a quantitative evaluation to compare the proposed semi-sparsity model with the aforementioned traditional methods. Notice that an objective evaluation is always not difficult for image decomposition methods because the ground truth of each component (structure or textures) is not available in practice. 

\begin{table*}[!htb]
    \small
    \caption{The correlation coefficient of STR metrics $C_0$/$C_1$ between image structure $u$ /$\nabla u$ (gradient) and image texture $v\!=\!f\!-\!u$ on the new dataset, and the computation time (s) compared with the state-of-the-art methods.}
    \label{Tab:tab3}
    \begin{center}
        \setlength{\tabcolsep}{0.1mm}{
        \begin{tabular}{|c|c|c|c|c|c|c|c|c|c|c|c|} 
            \hline
            \makecell{Metrics\\STR(19.23)}  & ROF\cite{rudin1992nonlinear}& $\mathrm{TV}$-$L^1$\cite{le2014cartoon+} & $\mathrm{TV}$-$G$\cite{meyer2001oscillating}& $\mathrm{TV}$-H$^{-1}$\cite{osher2003image}& $\mathrm{TV}$-$G$-$\mathcal{H}$\cite{xu2022new}& BTF\cite{sun2017image}& RTV\cite{xu2012structure}& $\mathrm{TGV}$-$L^1$\cite{bredies2013properties}& $\mathrm{TGV}$-$\mathcal{H}$\cite{jung2015simultaneous}& HTV-$\mathcal{H}$\cite{jung2015simultaneous} & Ours\\ 
            \hline
            \hline
            \makecell{$C_0(\downarrow)$}    & 0.1250    & 0.0464    & 0.0849    & 0.0462    & 0.0774    & 0.1438  &\underline{0.0381}   & 0.0850    & 0.0666   & 0.0742   & \textbf{0.0361}\\ 
            \makecell{$C_1(\downarrow)$}    & 0.0528    & \underline{0.0393}    & 0.0400    & 0.0467     & 0.0779    & 0.1018  & 0.0442              & 0.0987    & 0.0740   & 0.1041   & \textbf{0.0239}\\          
            \hline
            \makecell{Time (s)}  &  5.91  &  6.31   & 11.46   & 10.82  & 16.25     & \underline{3.61} & \textbf{2.05}  & 18.20  & 22.61  & 15.52  & 12.94\\
            \hline
    \end{tabular}}
    \end{center}
\end{table*}

As suggested in the work~\cite{jung2015simultaneous, aujol2006structure}, the structural and textural components show very little correlation for many texture-rich images. Based on this idea, the correlation coefficient of image structures and textures is then employed to evaluate the quality of decomposed results.  We here verify their relationship in Fig. \ref{Fig:fig8} (a)$\sim$(c). As we can see, the preferred image structures and textures indeed reveal very different characteristics in local regions. The relationship is more clear when comparing the textures $v$ with the gradient map of structures $\nabla u$ as indicated in Fig. \ref{Fig:fig8} (c)$\sim$(d). Specifically, we show their overlap regions in Fig. \ref{Fig:fig8} (e) in \textcolor{blue}{Blue}, where the former textures $v$ with the gradient map of structures $\nabla u$ are shown in \textcolor{red}{Red} and \textcolor{green}{Green}. The conclusion can be also understood from the characteristics of image structures and textures discussed in Sec. \ref{subsec:observation}. Accordingly, the strong edges and oscillating patterns are attributed to different parts --- the former belongs to image structures and the latter occurs frequently in image textures. As a result, image structures or the corresponding gradient --- referred as a representation of strong edges, has little correlation with image textures, since they have tiny overlap in spatial. 

The observation also motivates us to use the correlation coefficient for quantitative evaluation. We here define the correlation coefficient $C(x,y)\!=\!\frac{cov(x, y)}{\sqrt{var(x) var(y)}}$ for variables $x$ and $y$, where $cov(\cdot)$ and $var(\cdot)$ refer to the covariance and variance of the counterparts, respectively. Specifically, we denote $C_0(u,v)$ as the correlation coefficient of image structures $u$ and textures $v$, and $C_1(\nabla u, v)$ as that of the gradient of image structures $\nabla u$ and textures $v$. Both of them are adopted to indicate the correlation between image structures and textures. For color images, each metric is computed and averaged by channel. In general, $C_1$ is smaller than $C_0$ due to the less correlation of the gradient $\nabla u$ and structure $v$.  We here evaluate our semi-sparsity model against the compared methods on the new dataset. For fairness, all parameters in each method are fine-tuned by hand to give similar smoothing structures, quantified by the STR index. The correlation coefficients $C_0$ and $C_1$ are listed in Tab. \ref{Tab:tab3}, which are evaluated for each method under the average structure-to-texture ratio  $STR=19.23$ for 32 images. As we can see, the proposed semi-sparse method achieves the best results in both $C_0$ and $C_1$ cases.  

In addition, we also compare the running time in Tab. \ref{Tab:tab3} to show the efficiency of each method. To evaluate the performance of each method thoroughly, the BTF~\cite{sun2017image} and RTV~\cite{xu2012structure} methods are adopted from the official implementations, and the $\mathrm{TV}$-$L^1$~\cite{le2014cartoon+} and $\mathrm{TGV}$-$L^1$~\cite{bredies2013properties} algorithms are carried out based on the primal-dual algorithm\cite{chambolle2011first, bredies2020higher}. The numerical solutions for other methods are based on a similar ADMM procedure in \textbf{Algorithm} \ref{Alg:alg1} with 100 iterations. Notice that the BTF~\cite{sun2017image} has the computational cost in several seconds depending on the filter kernel size. The RTV~\cite{xu2012structure} is much faster for small-size images, while the time rises greatly due to the high computational cost of linear solver in large-scale cases. The primal-dual algorithm and ADMM procedure have a similar level of computation in each iteration since they are all determined by the FFT implementation in linear system solvers. The statistical results Tab. \ref{Tab:tab3} are estimated by processing a $512\times512$ resolution color image. All the methods are implemented in Matlab 2015b without any optimization on a desktop PC with Intel Core i7-9800X CPU 3.80GHz and 64G RAM.

\begin{figure}[!t]
	\begin{center}
		\subfloat[Input]
		{\includegraphics[width = 0.16\textwidth]{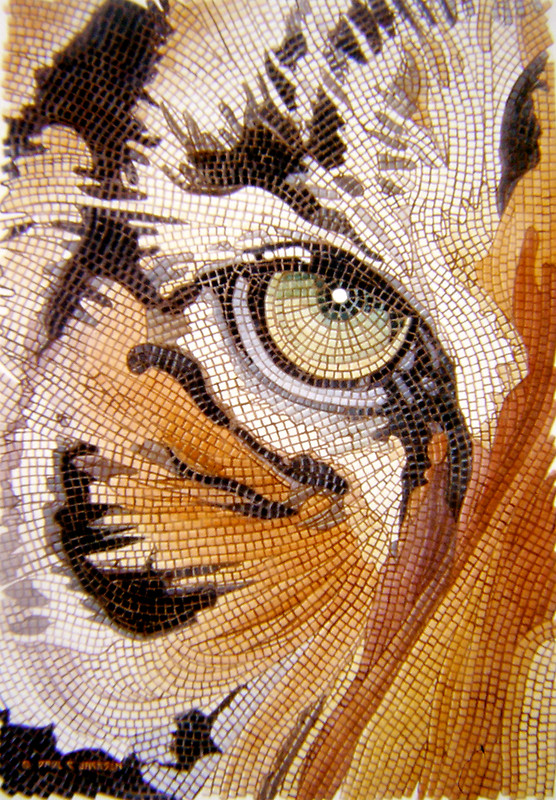}}\hfil
		\subfloat[$2^{nd}$-order]
		{\includegraphics[width = 0.16\textwidth]{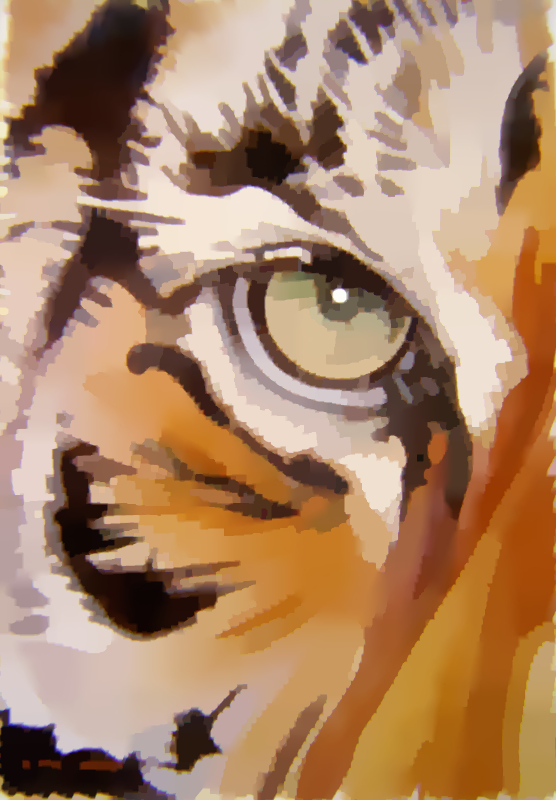}}\hfil
		\subfloat[$3^{rd}$-order]
		{\includegraphics[width = 0.16\textwidth]{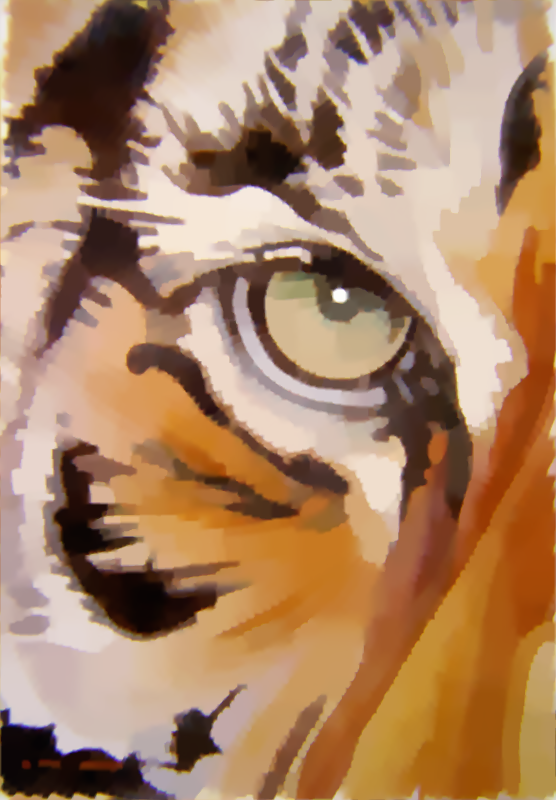}}\hfil
	\end{center}
	\caption{Visual comparison of image structural results with $L_0$-norm regularization on the (a) second-order ($\text{STR}\!=\!18.94$), and third-order ($\text{STR}\!=\!18.82$) gradient domains, respectively.} 
 \label{Fig:fig9}
\end{figure}

\section {Extensions and Analysis}
\label{extensions_and_analysis}

There are some obvious extensions for our semi-sparsity image decomposition model. Here, we briefly discuss two potential directions: applying the higher-order (order $n\ge 3$) $L_0$ regularization for semi-sparsity priors, and replacing the textural analysis model in other spaces.

\subsection{Higher-order Regularization} 

In the previous, we have discussed the semi-sparsity model with $L_0$-norm regularization in the second-order ($n\!=\!2$) gradient-domain in Eq. \ref{Eq:eq4} and \ref{Eq:eq5}, while it is straightforward to apply it in the higher-order ($n\ge3$) cases. In general, the choice $n$ is determined by the characteristics of image signals. As illustrated in semi-sparsity smoothing filter~\cite{huang2023semi}, it is usually enough to produce acceptable results by setting $n\!=\!2$ for natural images, although a better result could be attained when the regularization is imposed on a higher-order gradient domain. The conclusion is derived from the statistic distribution of higher-order gradients of natural images. In the context of image structure and texture decomposition, we claim that this conclusion is still valid for the similar propriety of image structures. Actually, the structural component of an image tends to more sparse in comparison of images when the textural part is decoupled or removed from images. 
\begin{figure*}[!htb]
	\begin{center}
		{\includegraphics[width=0.19\textwidth]{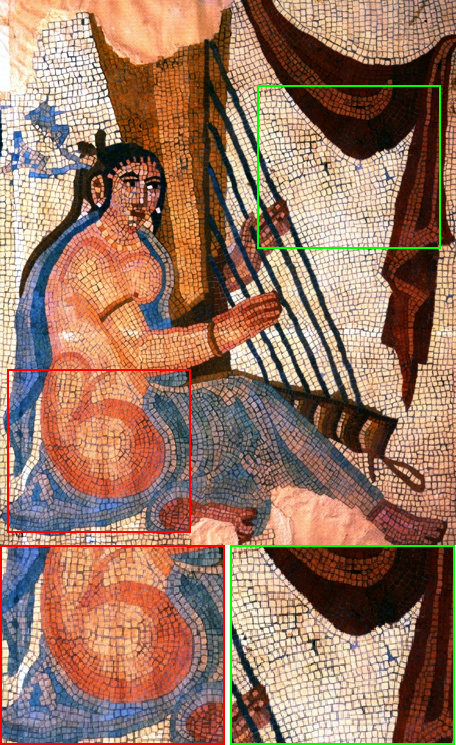}}\hfil
		{\includegraphics[width=0.19\textwidth]{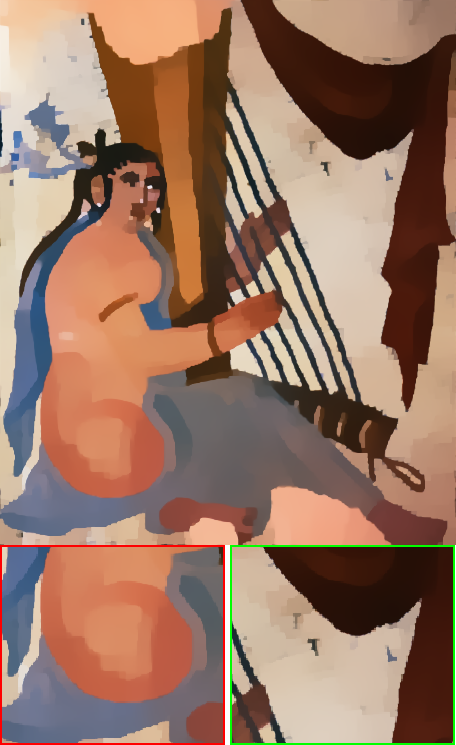}}\hfil
		{\includegraphics[width=0.19\textwidth]{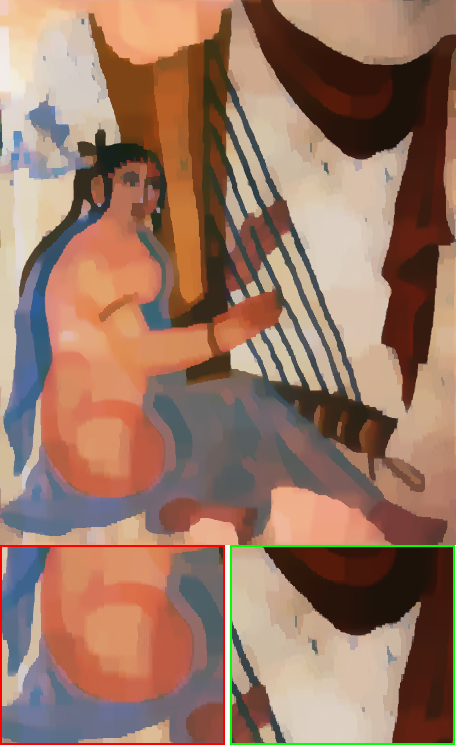}}\hfil
		{\includegraphics[width=0.19\textwidth]{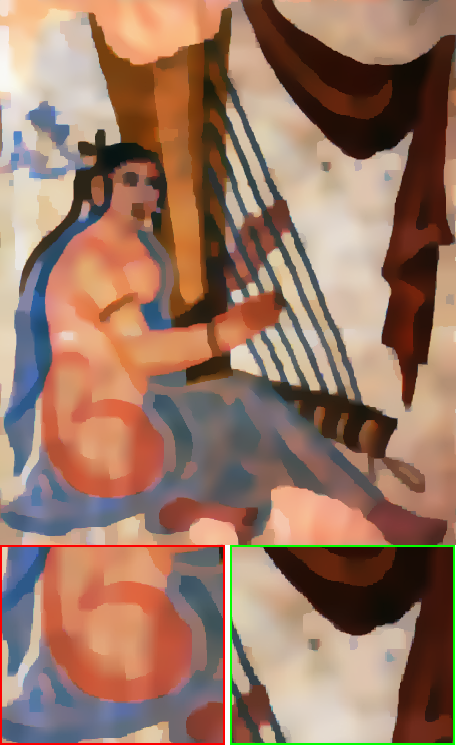}}\hfil
		{\includegraphics[width=0.19\textwidth]{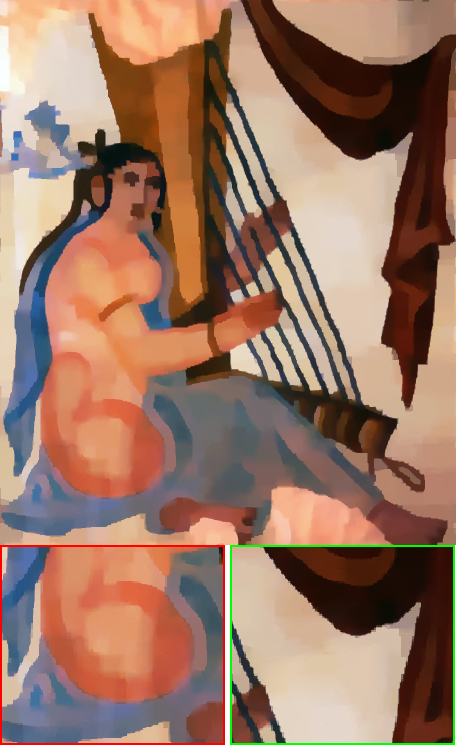}}\hfil
            \subfloat[Input]
		{\includegraphics[width=0.19\textwidth]{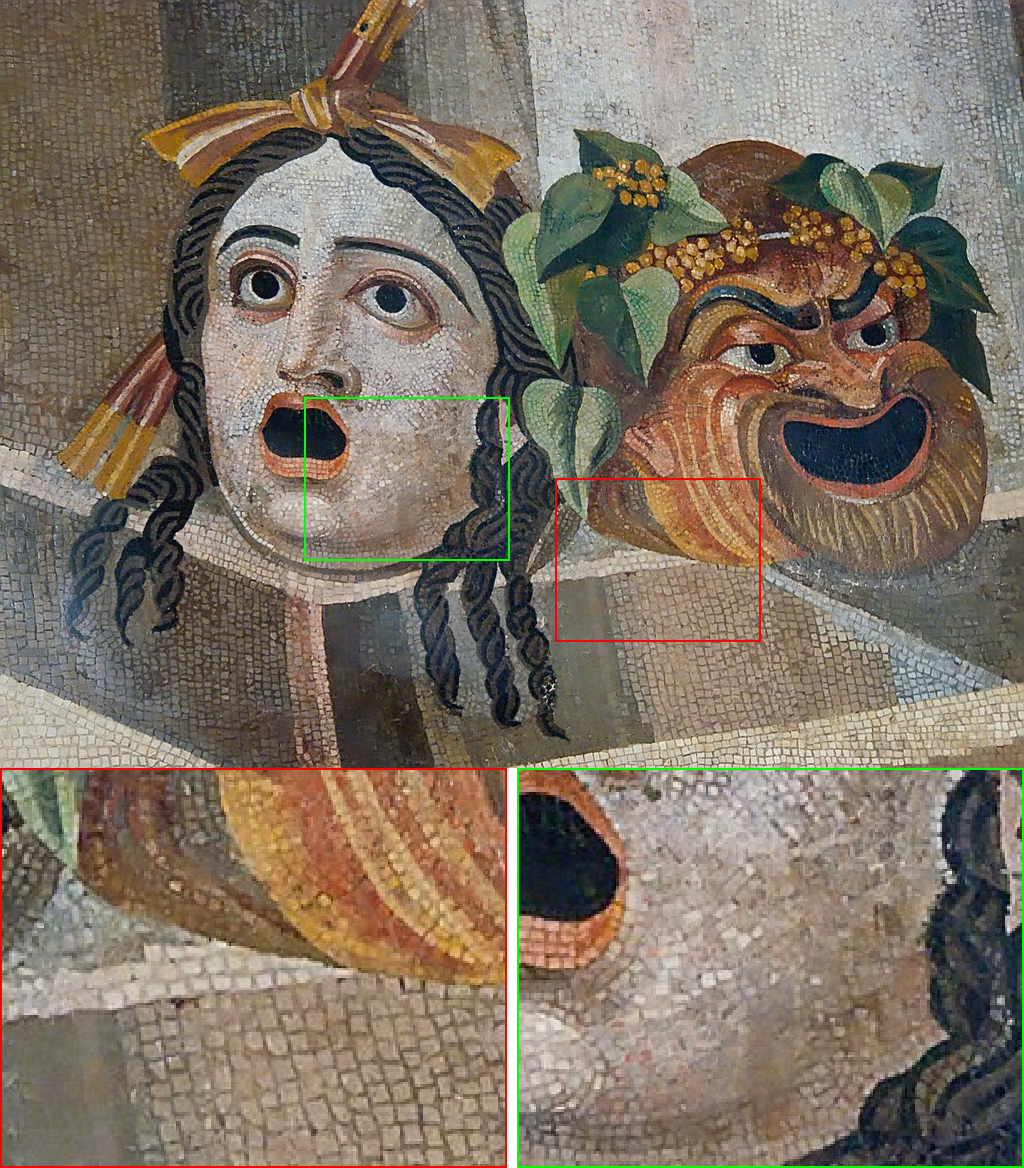}}\hfil
		\subfloat[RTV~\cite{rudin1992nonlinear}]
		{\includegraphics[width=0.19\textwidth]{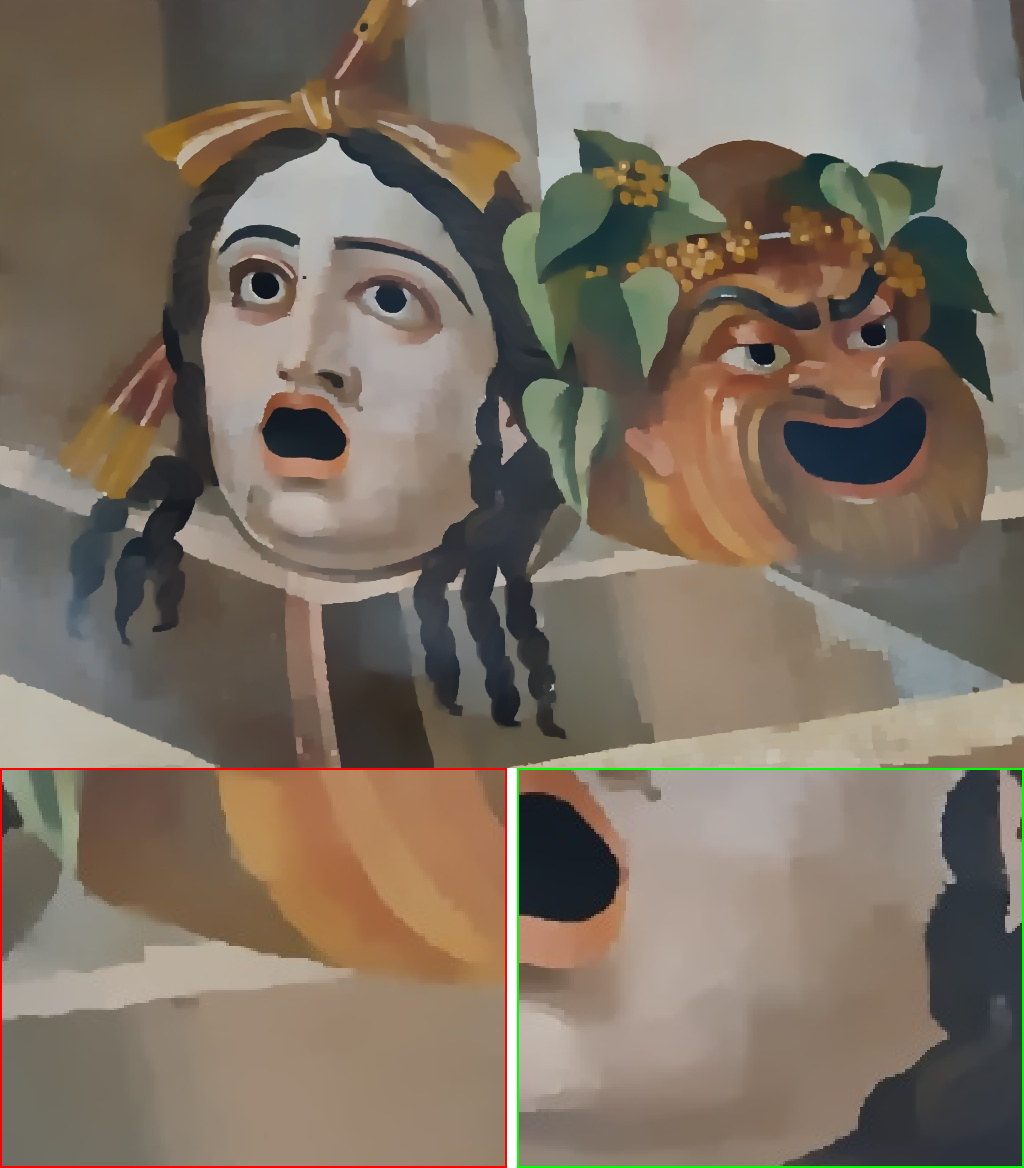}}\hfil
		\subfloat[$G$ space (Ours)]
		{\includegraphics[width=0.19\textwidth]{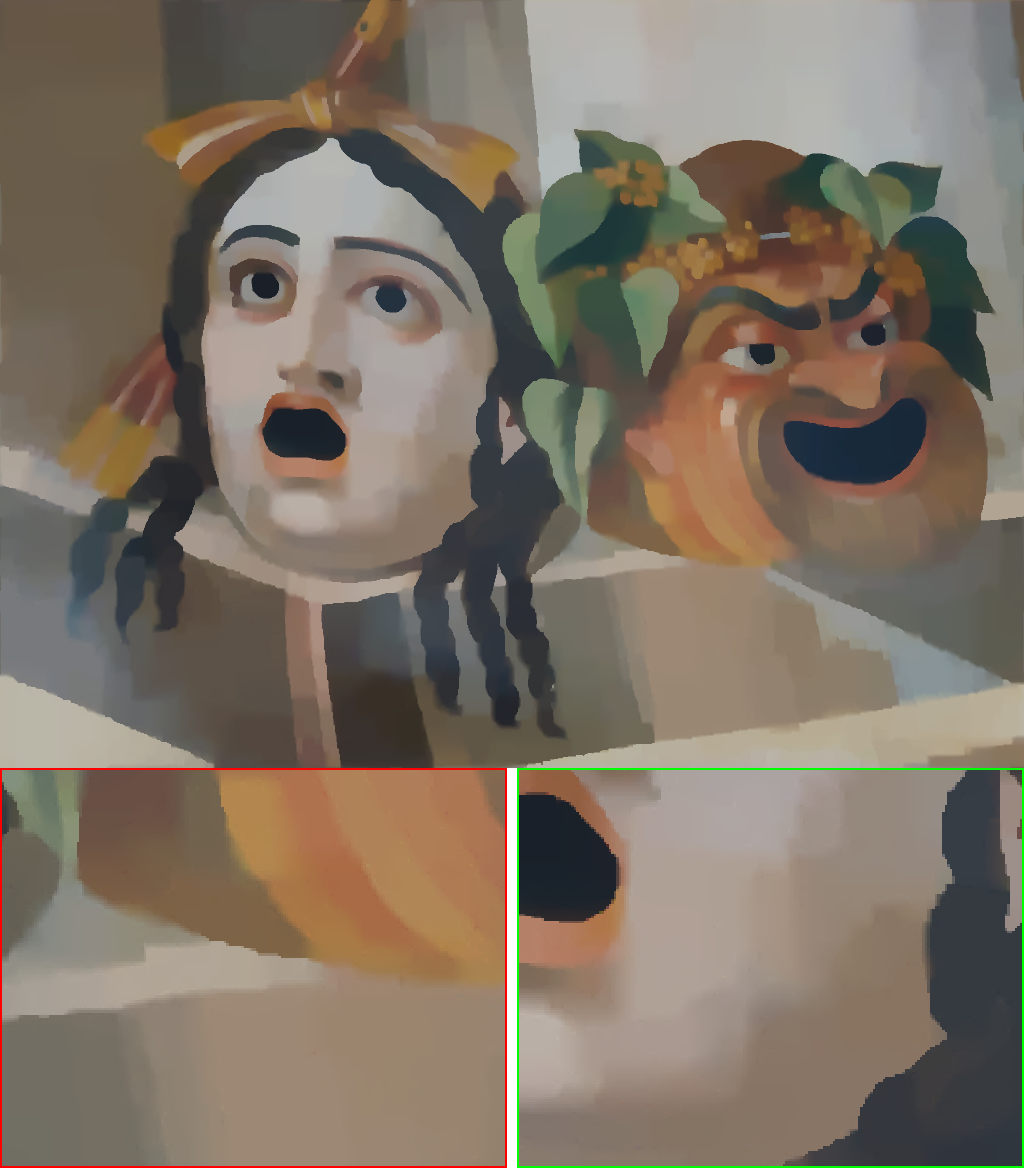}}\hfil
		\subfloat[$H^{-1}$ space (Ours)]
		{\includegraphics[width=0.19\textwidth]{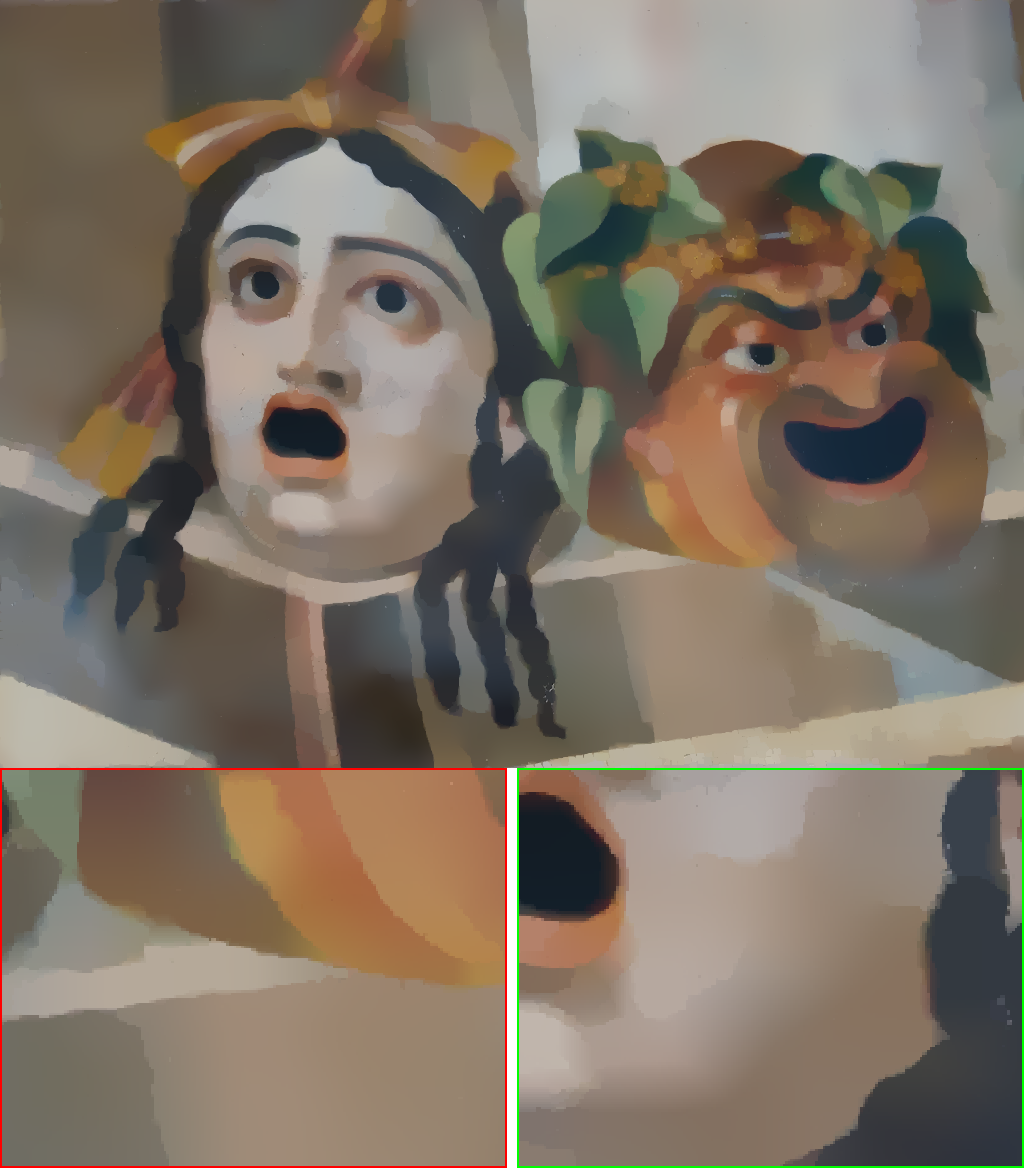}}\hfil
		\subfloat[$L^1$ space (Ours)]
		{\includegraphics[width=0.19\textwidth]{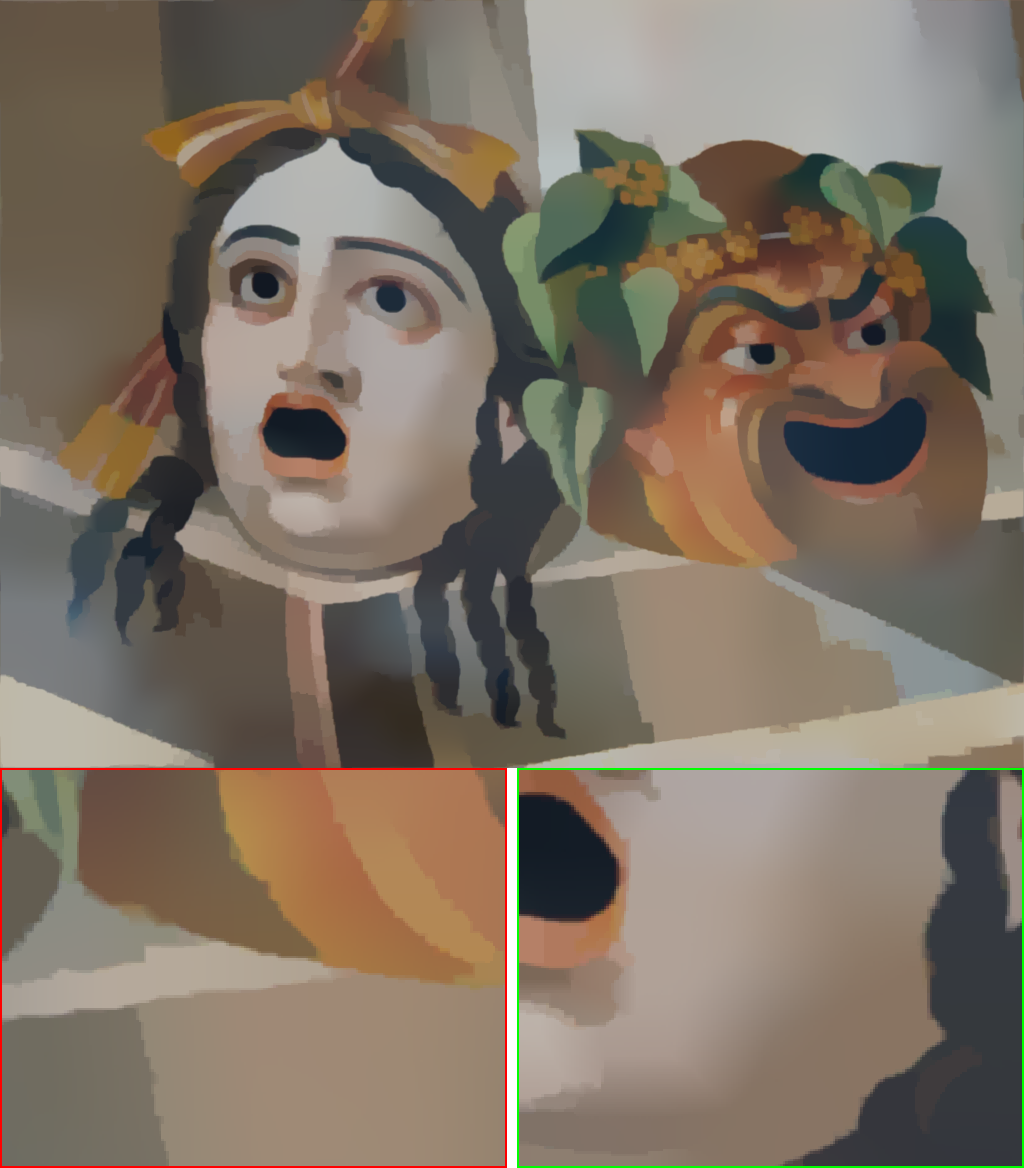}}\hfil
	\end{center}
	\caption{Visual comparison of structural decomposition results based on our semi-sparse model with the textural models in different spaces. (a) Input,  (b) RTV~\cite{xu2012structure},  (c) $G$ space, (d) $H^{-1}$ space, and (e) $L^1$ space. The structures are compared under STR metrics: 19.18 (top) and 22.10 (bottom), respectively.}
    \label{Fig:fig10} 
\end{figure*} %

We have shown the results in different scenarios of image structures and textures by using second-order $L_0$-norm regularization. For complementary, we compare the results by imposing the $L_0$ constraint on the third-order gradient domain. As shown in Fig. \ref{Fig:fig9}, the given image tends to have piece-wise constant and smoothing structures. In both cases, image textures are obviously removed despite that the tiny difference in the piece-wise smoothing surfaces. This is also indicated that it is usually precise enough to use second-order $L_0$-norm regularization. This conclusion is also demonstrated in other higher-order methods such as the well-known TGV-based methods~\cite{bredies2010total} and $\mathrm{TV}\text{-}\mathrm{TV}^2$ regularized models~\cite{bergounioux2010second, papafitsoros2014combined}. Another consideration of choosing the $2^{nd}$ order gradient for regularization is to reduce the computation cost in many practical applications.

\subsection{Textural Analysis in Different Spaces} 

In this paper, we have demonstrated the effectiveness of the higher-order $L_0$-norm regularization with the textural analysis in $L^1$ space. Apparently, it is possible to combine such a regularization with different textural models. 

As shown in Tab. \ref{Tab:tab2}, we have discussed several textural models that can be integrated with our structural analysis. The weak space $G$ \cite{meyer2001oscillating}, for instance, allows capturing the oscillatory components such as regular textures and random noise. The original space $G$ is equipped with the norm: 
$$
\|v\|_G=\inf _{\mathrm{g}=\left(g_1, g_2\right)}\left\{\left\|\sqrt{g_1^2+g_2^2}\right\|_{L^{\infty}} \mid v=\operatorname{div} g\right\},
$$
where $\operatorname{div}$ is the divergence operator. However, the direct numerical solution is unavailable for the involved $L^{\infty}$ space. The problem is then relaxed, for example, by replacing $G$ space with $G_p = W^{-1,p} (1\le p<\infty)$. By analogy, we combine the semi-sparse regularization with the $G_p$ textural analysis, giving a new semi-sparsity decomposition model,
\begin{equation}
\begin{aligned}
\mathop{\min}_{u,g} \lambda {\left\Vert u\!+\!\operatorname{div} g\!-\!f\right\Vert }_2^2\!+\! \gamma{\left\Vert g\right\Vert }_p^p \!+\! \alpha  {\left\Vert {\nabla}u\right\Vert }_1\!+\!\beta{\left\Vert \nabla^2 u \right\Vert}_0
\end{aligned}
\label{Eq:eq19}
\end{equation}
where $u = \operatorname{div} g$ with $g$ in $L^p$ space, and $\lambda, \gamma$ are positive weights. The difference of Eq. \ref{Eq:eq19} from the original OSV model~\cite{vese2003modeling} is the utilization of semi-sparsity regularization for structural analysis.

Another well-known structural analysis space is based on the different specializations of Hilbert space $\mathcal{H}$. As stated in~\cite{aujol2006structure}, there are several models can be deduced from  $\mathcal{H}$ space, including ROF model ($\mathcal{H}\!=\!L^2$)~\cite{rudin1992nonlinear}, $\mathrm{TV}$-$H^{-1}$ ($\mathcal{H}\!=\!H^{-1}$)~\cite{osher2003image} and $\mathrm{TV}$-Gabor method~\cite{liu2018new}. By analogy, it is straightforward to combine our semi-sparsity structural regularization with them for  preferable results. We introduce a simple modification of the $\mathrm{TV}$-$H^{-1}$ model~\cite{osher2003image}, 
\begin{equation}
\begin{aligned}
\mathop{\min}_{u,g} \lambda {\left\Vert f\!-\!u\right\Vert}_{H^{-1}}^2 \!+\! \alpha {\left\Vert{\nabla}u\right\Vert }_1\!+\!\beta{\left\Vert \nabla^2 u \right\Vert}_0
\end{aligned}
\label{Eq:eq20}
\end{equation}
where $\|f-u\|_{H^{-1}}^2\!=\!\int \left|\nabla\left(\Delta^{-1}\right)(f\!-\!u)\right|^2 dxdy$ for 2D cases. In the $\mathrm{TV}$-$H^{-1}$ model, the oscillatory texture is modeled as the second derivative of a function in a homogeneous Sobolev space, but the underlying smooth structure may be not handled well in view of the property of $\mathrm{TV }$ regularization, while our semi-sparsity extension in higher-order gradient domains partially avoids this problem.

For a supplement, we compare the modified models of Eq. \ref{Eq:eq19} and \ref{Eq:eq20} with the original one in $L^1$ space. The numerical solutions are also based on the similar multi-block ADMM procedure in \textbf{Algorithm} \ref{Alg:alg1}. As shown in Fig. \ref{Fig:fig10}, we take two typical images into account and both of them have large  oscillating textures intertwined with piece-wise constant and smoothing structures. Clearly, the textural models in weaker space $G$ and $H^{-1}$ tend to give similar results as the one in $L^1$ space when combined with the semi-sparsity higher-order $L_0$ regularization,  while the one in $H^{-1}$ space reveals slight degradation in local structural regions. An advantage of these modified models is that they tend to preserve more local structural information when compared with the cutting-edge RTV method in a similar level of smoothness. Notice also that the textural model in $L^1$ space generally produces better visual results with more sharpening edges among the piece-wise smoothing surfaces. This may be due to the fact that the weaker space $G$ and $H^{-1}$ tend to capture the periodical oscillating details, while the underlying image structures and textures could be in very complicated forms. As a result, they may produce unexpected results around the irregular textural parts, for example, causing blur edges nearby the non-periodical textures. In contrast, the model in $L^1$ space is more robust to these complex situations and this is also one of the reasons that we model the textural part in $L^1$ space. The conclusion is also verified in Tab.  \ref{Tab:tab4}, where the textural model in $L^1$ space has slightly better performance in both quality and efficiency. Finally, we would like to mention that the semi-sparsity higher-order regularization can be also utilized for more complex image decomposition problems, for example, the situation of structure, texture, and noise. More exploration and discussion are out of scope here. The reader is also referred to the work~\cite{shi2021image,jung2015simultaneous,lieu2008image,aujol2006structure, meyer2001oscillating} for more related details. 

\begin{table}[!t]
    \small
    \caption{The quantitative comparison of the STR metrics $C_0$/$C_1$ and running time for the textural models in different spaces.}
    \label{Tab:tab4}
    \begin{center}
        \setlength{\tabcolsep}{1mm}{
        \begin{tabular}{|c|c|c|c|c|} 
            \hline
            \makecell{Metrics\\STR(19.23)}  & RTV\cite{xu2012structure}&  $G$ space& $H^{-1}$ space & $L^1$ space\\ 
            \hline
            \hline
            \makecell{$C_0(\downarrow)$}   & 0.1444    &0.1287    &0.2145    &\textbf{0.0874}\\       
            \makecell{$C_1(\downarrow)$}   & 0.1174    &0.0966    &0.1383    &\textbf{0.0826}\\  
            \hline
            \makecell{Time (s)}         &\textbf{2.11}  &29.52  &26.83  &25.92\\
            \hline
    \end{tabular}}
    \end{center}
\end{table}

\section{Conclusion}
\label{sec:conclusion}

In this work, we have proposed a simple but effective semi-sparsity model for image structure and texture decomposition. We demonstrate its advantages based on a fast ADMM solver. Experimental results also show that such a semi-sparse minimization has the capacity of preserving sharp edges without introducing the notorious staircase artifacts in piece-wise smoothing regions and is also applicable for decomposing image textures with strong oscillatory patterns when applied to natural images. Some avenues of research for more complex decomposition models are also possible based on the semi-sparsity priors, which are leaving for further work.



\vspace{-6mm}
\begin{IEEEbiography}[{\includegraphics[width=1in,height=1.25in,clip,keepaspectratio]{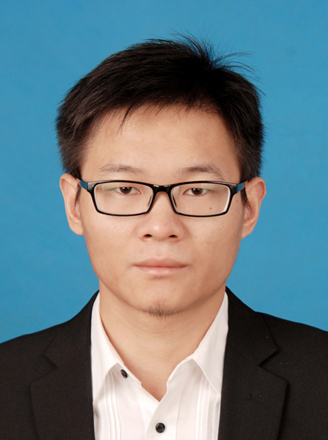}}]{Junqing Huang}
	received the BS degree in Automation from the School of Electrical Engineering, Zhengzhou University, Zhengzhou, China, in 2011, and the MS degree in Mathematics from the School of Mathematical Sciences, Beihang University (BUAA), Beijing, China, in 2015. He is currently a Ph.D. candidate of Department of Mathematics: Analysis, Logic and Discrete Mathematics, Ghent University, Belgium. His research interests include deep learning, image processing, optimal transport and optimization.
\end{IEEEbiography}

\vspace{-6mm}
\begin{IEEEbiography}[{\includegraphics[width=1in,height=1.25in,clip,keepaspectratio]{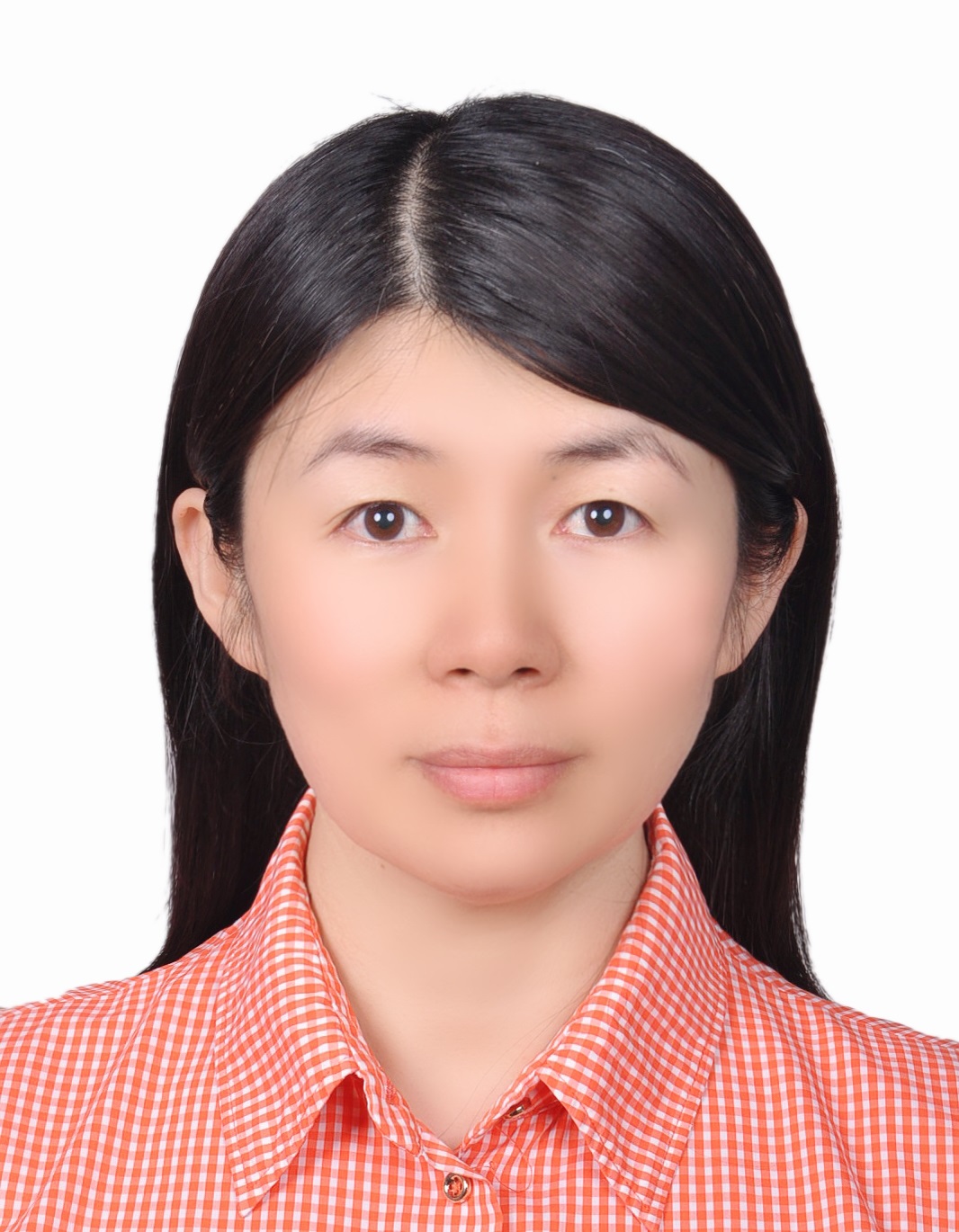}}]{Haihui Wang}	
	received the PhD degree in mathematics from Beijing University, Beijing, China, in 2003. She is currently a 
 full professor with the school of Mathematics and Sciences, Beihang University (BUAA), Beijing, China. Her research interests include artificial intelligence, machine learning, signal and image processing, wavelet analysis and applications.
\end{IEEEbiography}

\vspace{-6mm}
\begin{IEEEbiography}[{\includegraphics[width=1in,height=1.25in,clip,keepaspectratio]{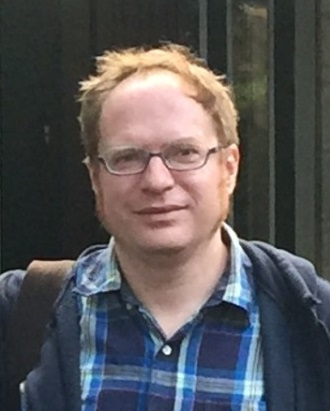}}]{Michael Ruzhansky} is currently a senior full professor in the Department of Mathematics and a
	Professorship in Special Research Fund (BOF) at Ghent University of Belgium, a Professorship in School of Mathematical Sciences at Queen Mary University of London, UK, and Honorary Professorship in Department of Mathematics at Imperial College London, UK.
	He was awarded by FWO (Belgium) the prestigious Odysseus 1 Project in 2018, he was recipient of several Prizes and Awards: ISAAC Award in 2007,
	Daiwa Adrian Prize in 2010 and Ferran Sunyer I Balaguer Prizes in 2014 and 2018. His research interests include different areas of analysis, in particular, theory of PDEs, microlocal analysis, and harmonic analysis.
\end{IEEEbiography}
\vfill

\end{document}